\pdfoutput=1
\documentclass[11pt]{article}

\usepackage[]{ACL2023}

\usepackage{times}
\usepackage{latexsym}
\usepackage{lipsum}
\usepackage{cleveref}

\usepackage{amssymb}% http://ctan.org/pkg/amssymb
\usepackage{pifont}% http://ctan.org/pkg/pifont

\usepackage{adjustbox}
\usepackage{multicol}
\usepackage{multirow}

\usepackage{graphicx}
\usepackage{microtype}
\usepackage{footmisc}
\usepackage{ctable}
\usepackage[normalem]{ulem}
\usepackage{longtable}
\usepackage{pdflscape}

\usepackage{CJKutf8}
\usepackage{array}
\newcommand*{\mline}[1]{%
\begingroup
   \begin{tabular}[c]{@{}>{\raggedright\arraybackslash}p{\linewidth}@{}@{}}#1\end{tabular}%
  \endgroup
}

\newcommand{\xmark}{\ding{55}}%

\usepackage[labelfont=bf]{caption}
\usepackage{subcaption}
\usepackage{mathrsfs}
\usepackage{bbding}
\usepackage{pifont}
\usepackage{soul}

\usepackage[T1]{fontenc}
\usepackage[utf8]{inputenc}

\usepackage{microtype}

\usepackage{inconsolata}
\usepackage{algorithm}
\usepackage{algpseudocode}
\usepackage{amssymb}
\usepackage{xurl}

\definecolor{greencustom}{HTML}{008000}

\newcommand{\error}[1]{\textcolor{red}{#1}}
\newcommand{\valid}[1]{\textcolor{greencustom}{#1}}

\title{A Multitask, Multilingual, Multimodal Evaluation of ChatGPT \\on Reasoning, Hallucination, and Interactivity}

\author{Yejin Bang\thanks{ \quad~Equal Contribution.}\quad~Samuel Cahyawijaya\quad~Nayeon Lee\quad~Wenliang Dai\quad~Dan Su\quad~Bryan Wilie\\
\textbf{Holy Lovenia\quad~Ziwei Ji\quad~Tiezheng Yu\quad~Willy Chung\quad~Quyet V. Do\quad~Yan Xu\quad~Pascale Fung$^*$}\\\
Centre for Artificial Intelligence Research (CAiRE)\\The Hong Kong University of Science and Technology\\
\texttt{yjbang@connect.ust.hk, pascale@ece.ust.hk}
}

\begin{document}
\maketitle
\begin{abstract}
 This paper proposes a framework for quantitatively evaluating interactive LLMs such as ChatGPT using publicly available data sets, using 23 data sets covering 8 different common NLP application tasks. We extensively evaluate the multitask, multilingual, and multi-modal aspects of ChatGPT based on these data sets and a newly designed multimodal dataset. We find that ChatGPT outperforms LLMs with zero-shot learning on most tasks and even outperforms fine-tuned models on some tasks. We find that it is better at understanding non-Latin script languages than generating them. It is able to generate multimodal content from textual prompts via an intermediate code generation step. Moreover, we find that ChatGPT is 63.41\% accurate on average in 10 different reasoning categories under logical reasoning, non-textual reasoning, and commonsense reasoning, hence making it an unreliable reasoner. 
ChatGPT suffers from hallucination problems like other LLMs.
Finally, the interactive feature of ChatGPT enables human collaboration with the underlying LLM to improve its performance, i.e., \textbf{8\% ROUGE-1} on summarization and \textbf{2\% ChrF++} on machine translation, in a multi-turn "prompt engineering" fashion. We release a code for evaluation set extraction.\footnote{\url{https://github.com/HLTCHKUST/chatgpt-evaluation}}

% This paper proposes a framework for quantitatively evaluating interactive LLMs such as ChatGPT using publicly available data sets. We carry out an extensive technical evaluation of ChatGPT using 23 data sets covering 8 different common NLP application tasks. We evaluate the multitask, multilingual and multi-modal aspects of ChatGPT based on these data sets and a newly designed multimodal dataset. We find that ChatGPT outperforms LLMs with zero-shot learning on most tasks and even outperforms fine-tuned models on some tasks. We find that it is better at understanding non-Latin script languages than generating them. It is able to generate multimodal content from textual prompts, via an intermediate code generation step. Moreover, we find that ChatGPT is 63.41\% accurate on average in 10 different reasoning categories under logical reasoning, non-textual reasoning, and commonsense reasoning, hence making it an unreliable reasoner. It is, for example, better at deductive than inductive reasoning. 
% ChatGPT suffers from hallucination problems like other LLMs and it generates more extrinsic hallucinations from its parametric memory as it does not have access to an external knowledge base. 
% Finally, the interactive feature of ChatGPT enables human collaboration with the underlying LLM to improve its performance, i.e, \textbf{8\% ROUGE-1} on summarization and \textbf{2\% ChrF++} on machine translation, in a multi-turn "prompt engineering" fashion. We also release codebase for evaluation set extraction.
%\footnote{\url{https://github.com/HLTCHKUST/chatgpt-evaluation}}

\end{abstract}

\section{Introduction}

% % Intro Large language models
% % ChatGPT
% Recently, ChatGPT\footnote{The first version was available public through online UI on November 2022.}, an LLM that enables user interaction in the form of conversation, has been released to the public. 
% \footnote{The first version was available public through online UI on November 2022.} 
% \footnote{\url{https://beta.openai.com/docs/model-index-for-researchers}} 
ChatGPT is a successor of the large language model (LLM) InstructGPT~\cite{instructGPT} with a dialog interface that is fine-tuned using the Reinforcement Learning with Human Feedback (RLHF)~\cite{rlhf} approach. ChatGPT has gathered 100 million monthly active users in such a short period of time \cite{Hu_2023} and is being used by businesses and consumers alike for a myriad of mostly textual tasks. One reason for its unprecedented popularity is that ChatGPT, through its scale and via RLHF, has shown impressive abilities in many areas of NLP as well as emergent abilities. Another reason is that its dialog interface allows users to interact with the underlying LLM more effectively and efficiently via interactive chats that are akin to multi-turn prompting.

However, despite its powerful abilities, anecdotal reports on ChatGPT consistently showed remaining challenges - for example, it fails in some elementary mathematical~\cite{gilson2022well,goldberg2023some,mathchatgpt,choi2023chatgpt,davis2023mathematics} and commonsense reasoning tasks~\cite{guo2023close,davis2023mathematics}; it hallucinates with human-like fluency and eloquence on things that are not based on truth~\cite{shen2023chatgpt,thorp2023chatgpt,smith2023why}; and as a general-purpose language model trained from everything on the web, its language coverage is questionable~\cite{lu2022trip,jiao2023chatgpttranslate}. Consequently, it is not clear what people can or cannot use ChatGPT for despite its popularity.

% \pascale{I don't know if it is odd but I think it is important to point out that this was done very early and was the first 3rd party benchmarking of this kind}. \bang{i think it is not odd.}
Since OpenAI never published any benchmarking results on ChatGPT at the time, seeing this need, in February 2023, we proposed a comprehensive framework for quantitatively evaluating interactive LLMs such as ChatGPT 
% ~\footnote{All the experiments are conducted with the Dec. 15 version of ChatGPT.}
through standard public test sets on major NLP tasks such as question answering, reasoning, summarization, machine translation, sentiment analysis, language identification, task-oriented dialogue, 
and misinformation detection. We evaluate its multilingual performance as well as vision-language multimodal abilities. With additional experiments, we also quantitatively evaluated its primary limitations in \textit{reasoning} and \textit{hallucination}. In addition, we conducted experiments to test its \textit{multi-turn interactivity} as a means for better prompt engineering. We aimed to provide insights to users of ChatGPT on the strengths mentioned above and limitations, as well as how they can improve outcomes with interactivity. To the best of our knowledge, this is the first published benchmark of ChatGPT from a third party. More recently, the GPT-4 technical report~\cite{openai2023gpt4} published a number of human task benchmarks.
% \pascale{More recently, the technical report on GPT-4 did publish a number of human task benchmarks.}

The true scope of all emergent capabilities of generative models, including ChatGPT, is still unclear. Thus, any benchmarking exercise cannot be 100\% ``comprehensive'' in the scientific sense. We aim to show not just researchers but also users what ChatGPT can and cannot do by presenting interpretable benchmarking results in a zero-shot setting without access to APIs so that the general audience can replicate our evaluation with the test sets we have provided in a zero-shot setting. This version of ChatGPT is 15 December 2022.
% \pascale{This version of ChatGPT is ...}

The following are the major insights we have gained from the evaluations: 

\textbf{Multitask, Multimodal, and Multilingual: } 
For 9/13 NLP datasets, ChatGPT outperforms previous LLMs with zero-shot learning. It even outperforms fully fine-tuned task-specific LMs on 4 different tasks. In other cases, ChatGPT is on par or slightly lower than fully fine-tuned for specific NLP tasks;
ChatGPT fails to generalize to low-resource and extremely low-resource languages (e.g., Marathi, Sundanese, and Buginese). There is an overall performance degradation in low-resource languages, especially in non-Latin scripts in the case of translation; its weakness lies in generation rather than understanding part of the translation process;
ChatGPT enables a code intermediate medium to bridge vision and language, even though the multi-modality ability is still elementary compared to vision-language models.

% \begin{itemize}
%     \item For 9/13 NLP datasets, ChatGPT outperforms previous LLMs with zero-shot learning. It even outperforms fully fine-tuned task-specific LMs on 4 different tasks. In other cases, ChatGPT is on par or slightly lower than fully fine-tuned for specific NLP tasks; 
%     \item ChatGPT fails to generalize to low-resource and extremely low-resource languages (e.g., Marathi, Sundanese, and Buginese). There is an overall performance degradation in low-resource languages, especially in non-Latin scripts in the case of translation; its weakness lies in generation rather than understanding part of the translation process;
%     \item ChatGPT enables a code intermediate medium to bridge vision and language, even though the multi-modality ability is still elementary compared to vision-language models. 
% \end{itemize}

\textbf{Reasoning: }
We tested 10 different reasoning categories with 634 samples in total. Based on our experiments, ChatGPT shows more weakness in inductive reasoning than in deductive or abductive reasoning. ChatGPT also lacks spatial and mathematical reasoning while showing better temporal reasoning. Further, we found that ChatGPT is relatively better at commonsense reasoning than non-textual semantic reasoning. Finally, while ChatGPT shows acceptable performance in causal and analogical reasoning, it is bad at multi-hop reasoning capability, similar to other LLMs' weakness~\citep{ott2023thoughtsource}. 
 
\textbf{Hallucination: }
Similar to other LLMs \citep{radford2019language,muennighoff2022crosslingual,bigscience2022bloom}, ChatGPT suffers from the hallucination problem. It generates more extrinsic hallucinations -- factual statements that cannot be verified from the source.
% , from its parametric memory across all tasks since it does not possess the access to external knowledge bases.

\textbf{Interactivity: }
One of the primary differentiating factors of ChatGPT from its predecessors is its \textit{multi-turn dialog interactivity}. This enables ChatGPT to perform multiple tasks within a dialog session. There is also significant performance improvement (8\% ROUGE-1 on summarization and 2\% ChrF++ on low-resource machine translation) via multi-turn interactivity in various standard NLP tasks. This process is akin to prompt engineering with feedback from the system.

% \paragraph{Organization of This Paper:} We first provide an overview of ChatGPT and related work (\S\ref{sec:related}). Then, we provide evaluation results on ChatGPT on  various application test sets, on multilingual test sets, and on a new multimodal task in \S\ref{sec:mmm}. We then explore the three main strengths and weaknesses of ChatGPT, namely \textit{reasoning} (\S\ref{sec:reasoning}), \textit{hallucination} (\S\ref{sec:hallu}) and \textit{interactivity} (\S\ref{sec:interactive}) in the subsequent three sections. Finally, we discuss and give a conclusion on our findings of ChatGPT. 

% \bang{[BIG QUESTION]: do we need to explain chatgpthere? how about appendix?}
% [RELATED WORK HERE ORIGINAL]

\section{Multitask, Multilingual, and Multimodal Evaluations of ChatGPT}
\label{sec:mmm}

\subsection{Multitask Ability of ChatGPT}
\label{sec:evaluating-multitask-chatgpt}

\begin{table*}[!t]
    \centering
    \resizebox{0.95\linewidth}{!}{
    \begin{tabular}{c|c|c|c|c|c|c}
    \toprule
    \textbf{Tasks} & \textbf{Dataset} & \textbf{Metric} & \textbf{Reference} & \textbf{\begin{tabular}[c]{@{}c@{}}Fine-Tuned\\SOTA\end{tabular}} & \textbf{\begin{tabular}[c]{@{}c@{}}Zero-Shot\\SOTA\end{tabular}} & \textbf{ChatGPT} \\ \midrule    \textbf{Summarization} & CNN/DM & ROUGE-1 & \citet{lewis-etal-2020-bart} & 44.47 & 35.27$^*$ & 35.29 \\
     & SAMSum & ROUGE-1 & \citet{lewis-etal-2020-bart} & 47.28 & - & 35.29 \\ \midrule
    \multirow{2}{*}{\textbf{\begin{tabular}[c]{@{}c@{}}MT\\(XXX$\rightarrow$Eng)\end{tabular}}} & FLoRes-200 (HRL) & ChrF++ & \citet{nllb2022nlbb} & 63.5 & - & 58.64 \\
     & FLoRes-200 (LRL) & ChrF++ & \citet{nllb2022nlbb} & 54.9 & - & 27.75 \\ \midrule
     \multirow{2}{*}{\textbf{\begin{tabular}[c]{@{}c@{}}MT\\(Eng$\rightarrow$XXX)\end{tabular}}} & FLoRes-200 (HRL) & ChrF++ & \citet{nllb2022nlbb} & 54.4 & - & 51.12 \\
     & FLoRes-200 (LRL) & ChrF++ & \citet{nllb2022nlbb} & 41.9 & - & 21.57 \\ \midrule
    \multirow{4}{*}{\textbf{\begin{tabular}[c]{@{}c@{}}Sentiment \\Analysis\end{tabular}}} & NusaX - Eng & Macro F1 & \citet{winata2022nusax} & 92.6 & 61.5 & 83.24 \\
     & NusaX - Ind & Macro F1 & \citet{winata2022nusax} & 91.6 & 59.3 & 82.13 \\
     & NusaX - Jav & Macro F1 & \citet{winata2022nusax} & 84.2 & 55.7 & 79.64 \\
     & NusaX - Bug & Macro F1 & \citet{winata2022nusax} & 70.0 & 55.9 & 55.84 \\ \midrule
    \multirow{6}{*}{\textbf{\begin{tabular}[c]{@{}c@{}}Question\\Answering\end{tabular}}} & bAbI task (15 | 16) & Accuracy & \citet{weston2016qa} & 100 | 100 & - & 93.3 | 66.7 \\
     % & bAbI task 16 & Accuracy & \citet{weston2016qa} & 100 & - & 66.7 \\
     & EntailmentBank & Accuracy & \citet{clark2018arc} & 86.5 & 78.58 & 93.3 \\
     & CLUTRR & Accuracy & \citet{minervini2020diffproving} & 95.0 & 28.6 & 43.3 \\
     & StepGame (k=9 | k=1) & Accuracy & \citet{mirzaee-kordjamshidi-2022-transfer} & 48.4 | 98.7 & - & 23.3 | 63.3\\
     % & StepGame (k=1) & Accuracy & \citet{mirzaee-kordjamshidi-2022-transfer} & 98.7 & - & 63.3 \\
     & Pep-3k & AUC & \citet{porada-etal-2021-modeling} & 67.0 & - & 93.3 \\ \midrule
    \multirow{2}{*}{\textbf{\begin{tabular}[c]{@{}c@{}}Misinformation\\Detection\end{tabular}}} & COVID-Social & Accuracy  & \citet{lee-etal-2021-towards} & 77.7  & 50.0 & 73.3\\
     & COVID-Scientific & Accuracy  & \citet{lee-etal-2021-towards} & 74.7 & 71.1 & 92.0\\ \midrule
    \multirow{3}{*}{\textbf{\begin{tabular}[c]{@{}c@{}}Task-Oriented\\Dialogue\end{tabular}}} & MultiWOZ2.2 & JGA & \citet{Zhao2022DescriptionDrivenTD} & 60.6 & 46.7 & 24.4 \\
     & MultiWOZ2.2 & BLEU & \citet{nekvinda-dusek-2021-shades} & 19.1 & - & 5.65 \\      
     & MultiWOZ2.2 & Inform Rate & \citet{Yang_Li_Quan_2021} & 95.7 & - & 71.1 \\ \midrule
    \multirow{2}{*}{\textbf{\begin{tabular}[c]{@{}c@{}}Open-Domain\\KGD\end{tabular}}} & OpenDialKG & BLEU | ROUGE-L& \citet{ji2022rho} & 20.8 | 40.0 & 3.1 | 29.5 & 4.1 | 18.6\\
     % & OpenDialKG & ROUGE-L & \citet{ji2022rho} & 40.0 & 29.5 & 18.6 \\ 
     & OpenDialKG & FeQA & \citet{ji2022rho} & 48.0 & 23.0 & 15.0 \\ 
     \bottomrule
     
    \end{tabular}
    }

    \caption{Performance of ChatGPT compared to state-of-the-art fully-fine-tuned models (Fine-Tuned SOTA) and LLM in zero-shot settings (Zero-Shot SOTA). The referenced performances are evaluation results on full test sets, while the ChatGPT performances are computed on subsets of the corresponding dataset \textbf{using 30 to 200 data samples} for each task. For Machine Translation (MT) tasks, we follow the definitions of high-resource language (HRL) and low-resource language (LRL) from NLLB~\cite{nllb2022nlbb} and take subsets of languages to represent each group. JGA denotes joint goal accuracy. \*Average of performances for CNN and DM from~\citet{goyal2022news}. LMs in zero-shot settings are as follows. Summarization: InstructGPT, MT: NLLB-200, Sentiment Analysis: XLM-R LARGE, QA: ST-MoE-32B, ZeroQA, GPT-3, Misinformation Detection:  GPT-2, Task-Oriented Dialogue: D3ST, Open-Domain KGD: GPT-Jurassic-6B.}
    \label{tab:multitask-performance}
    \vspace{-0.8em}
\end{table*}

% HRL denotes high-resource language. LRL denotes low-resource. 

% \pascale{this section is extremely weak compared to the section on multiinguality and reasoning!!! You need to add more details.}\bang{Added}

ChatGPT has become very well-known in such a short period of time to general public users, not just those who are in AI, machine learning, and NLP communities who might be more familiar with LLMs. One of the main reasons is that, in addition to media reports, innumerable use cases of ChatGPT are shared by both non-academic and academic users online~\citep{marr_2022,gordon_2023,shankland_2023}. There have been debates and panels on whether ChatGPT is approaching Artificial General Intelligence, as it seems to be able to carry out a multitude of tasks without specific fine-tuning~\citep{desk_2023,johnson_2023,kingson_2023}. 
On the other hand, there has also been as much sharing of its failures in simple tasks \citep{gilson2022well,choi2023chatgpt,shen2023chatgpt}.
% \pascale{citation} 

% It is hard to grasp meaningful insights with a limited selection of examples, which could be possibly affected by selection bias as shown in its mathematical capability~\cite{mathchatgpt}.
% \pascale{WHAT? We are using a limited selection of samples from test sets!}

Instead of relying on anecdotal examples, we first evaluate ChatGPT's performance in various standard NLP 
tasks in a zero-shot manner to obtain a basic/better understanding of its multi-task ability. We compile results from the existing literature on ChatGPT and compare them with the state-of-the-art fully-fine-tuned and zero-shot models across multiple tasks. We evaluate ChatGPT performances on 21 datasets covering 8 tasks, i.e., summarization, machine translation, sentiment analysis, question answering, task-oriented dialogue, open-domain knowledge-grounded dialogue, and misinformation detection tasks. We sample testing cases from existing standard test sets for each task with a sample size ranging from 30 to 200 samples.

\paragraph{Multitask Generalization of ChatGPT} The result of the multitask evaluation is shown in~\Cref{tab:multitask-performance}. ChatGPT is shown to achieve remarkable zero-shot performances on multiple tasks, surpassing previous state-of-the-art zero-shot models on 9 out of 13 evaluation datasets with reported zero-shot LLMs' performances. In most tasks, especially task-oriented and knowledge-grounded dialogue tasks, task-specific fully-fine-tuned models outperform ChatGPT. Compared to the latter, ChatGPT yields lower performance in most tasks while still surpassing the performance on 4 datasets. 

Furthermore, from the evaluation results, we also observe several limitations of ChatGPT: 1) limited language understanding and generation capabilities on low-resource languages, 2) lacking reasoning ability as shown from the results in QA, and 3) performing task-oriented and knowledge-grounded dialogue tasks. More detailed experimental setup and analysis for each task are shared in Appendix \S\ref{appendix:multitask-details}. We also provide the complete list of all the datasets used in our evaluation in Appendix~\ref{sec:datasets}.

\paragraph{ChatGPT on Dialogue Tasks}
\label{sec:dialo}
Given that ChatGPT has the ability to generate conversation-like responses, we test it on conventional dialogue tasks: 1) knowledge-grounded open-domain dialogue and 2) Task-oriented dialogue. Task setups are explained in \Cref{appendix:dialogue}.

% it is interesting to test their ability in response generation in different dialogue settings
\paragraph{Knowledge-Grounded Open-Domain Dialogue}
% Open-domain dialogue systems interact with humans with generated responses automatically and aim to provide users with an engaging experience.
% To boost informativeness, these systems leverage external knowledge, including structured knowledge such as knowledge graphs~\cite{zhao2020knowledge,ji2022rho}  and unstructured knowledge such as free text~\cite{xu2022retrieval}. 
To quantitatively measure ChatGPT's performance on knowledge-grounded dialogue, we utilize 50 samples from the test set of OpenDialKG~\citep{moon2019opendialkg}, which contains open-ended dialogues grounded on a knowledge path.
% We use the following instruction for this KGD task: \texttt{``Can we try dialogue generation? I will give you turns, and you can generate the next turn, but only one.\textbackslash n \textbackslash n You can also consider the knowledge of XXX for your reference in the dialogue.''}
According to human judgment, the responses from ChatGPT are of high quality with fluent response generation and incorporating the provided knowledge in the response. However, the automatic evaluation results are relatively low compared with fine-tuned GPT2. We postulate this is because ChatGPT responses are longer than the golden answers and include content from its parametrized knowledge injected during pre-training.

% Specifically, ChatGPT obtains a 4.05 BLEU and an 18.62 ROUGE-L score as the generated responses tend to be longer than the golden answers.
% For FeQA, which measures the generated response's faithfulness to the input source, ChatGPT gets 15.03 since some generated responses include content from its parametrized knowledge injected during pre-training.

\paragraph{Task-Oriented Dialogue}
% In task-oriented dialogue (TOD), a model needs to fulfill a specific objective by interacting in natural language with the user. This task is often split into three modules: natural language understanding with belief state tracking, decision-making through dialogue policies, and response generation -- a modular approach that handles each of these steps with different models. Besides, unified approaches are starting to show increasingly strong performances \citep{hosseini2020simple,peng2021soloist}. 
% Although ChatGPT seems more appropriate for open-domain dialogue tasks, 
We investigate and discuss how ChatGPT's emergent abilities and interactivity could potentially be leveraged for ToD as well in two setups. Firstly, A) modular approach: testing dialogue state tracking (DST) and response generation using oracle actions. 
DST is mediocre while ChatGPT successfully leverages all information provided while answering the questions with a 71.1\% inform rate and 5.65 BLEU score. Next, B) Unified approach: a direct approach to simulate the ToD interaction while leveraging information in a structured database. 
We observed the limitations of ChatGPT: 1) ChatGPT cannot keep the belief state across multiple turns within the interaction, 2) ChatGPT's response tends to be wrong if the query introduces a basic level of reasoning 3) ChatGPT tends to generate hallucinated information beyond the given knowledge, which is not desirable for ToD.
We provide details and examples of the modular and unified approaches in Appendix~\ref{sec:multiturn-tod}.

% \begin{table}[!t]
% \centering
% \resizebox{1\linewidth}{!}{
% \begin{tabular}{lccc}
% \toprule
% \textbf{Model} &  \textbf{BLEU $\uparrow$} & \textbf{ROUGE-L $\uparrow$} & \begin{tabular}[c]{@{}c@{}}\textbf{FeQA $\uparrow$} \\ \cite{durmus-etal-2020-feqa}\end{tabular} \\ \midrule
% ChatGPT & 4.05	& 18.62 & 15.03 \\
% GPT2 & 11.10 & 30.00 & 26.54 \\
% \bottomrule
% \end{tabular}
% }
% % \quad\quad
% \caption{Automatic evaluation results on OpenDialKG. The results for GPT2 are from \citet{dziri-etal-2021-neural}.}
% \label{tab:chatgpt on OpenDialKG}
% % \vspace{-1em}
% \end{table}

% \begin{table}[]
% \centering
%     \begin{adjustbox}{width={0.36\textwidth},totalheight={\textheight},keepaspectratio}
% \begin{tabular}{c|cc}
% \toprule
% State Tracking & \multicolumn{2}{c}{Response Generation} \\  \cmidrule(l{2pt}r{2pt}){1-1} \cmidrule(l{2pt}r{2pt}){2-3}
% Joint Goal Acc.  & BLEU & Inform rate\\  \cmidrule(l{2pt}r{2pt}){1-1} \cmidrule(l{2pt}r{2pt}){2-3}
% 24.4\% & 5.65 & 71.1\% \\ \bottomrule
% \end{tabular}
% \end{adjustbox}
% \caption{Result for Task-oriented Dialogue Setup A -- Modular Approach.}
% \label{tab:tod}
    
% \end{table}

% # [MULTI TURN DETAILS WERE HERE --> APPENDIX ]

\subsection{Evaluating Multilinguality of ChatGPT}
\label{sec:language-disparity}

%\footnotetext{https://en.wikipedia.org/wiki/List_of_languages_by_total_number_of_speakers}

Training data size affects language understanding and generation ability of LMs~\citep{raffel2020t5,cahyawijaya2021indonlg,rae2021gopher,bigscience2022bloom,chowdhery2022palm,hoffmann2022chinchilla}. As an LLM, the same premise also applies to ChatGPT, but the question is to what extent. 
We investigate this question through a series of experiments by analyzing 1) the language understanding 
capability through sentiment analysis (SA) and language identification (LID) tasks, and 2) the language generation capability through machine translation using English as the pivot language. Based on the size proportion in CommonCrawl (i.e., the primary source of language pre-training data used in various LLMs)
% \footnote{CommonCrawl is the primary source of language pre-training data used in various LLMs}
, we group languages into 4 language resource categories, i.e., \textbf{high-resource language (HRL)} (<$\geq$1\%), \textbf{medium-resource language (MRL)} ($\geq$0.01\%), \textbf{low-resource language (LRL)} ($\geq$0.0001\%), and \textbf{extremely low-resource language (X-LRL)} (<0.0001\%). The statistics of the languages are shown in~\Cref{tab:cc_distribution} and other details are described in Appendix~\ref{app:multilinguality}.

% 5 categories, i.e., \textbf{extremely high-resource}, \textbf{high-resource}, \textbf{medium-resource}, \textbf{low-resource}, and \textbf{extremely low-resource}. For the LID experiment, we choose a set of 10 languages to be studied, i.e., English (EN), Chinese (ZH), French (FR), Indonesian (ID), Vietnamese (VI), Korean (KO), Hindi (HI), Marathi (MR), Javanese (JV), and Sundanese (SU). 
% we group the languages into 5 categories: \textbf{extremely high-resource} (EN), \textbf{high-resource} (ZH, FR), \textbf{medium-resource} (ID, VI, KO), \textbf{low-resource} (HI, MR), and \textbf{extremely low-resource} (JV, SU).

\begin{table}[!t]
    \centering
    
    \resizebox{0.8\linewidth}{!}{
    \begin{tabular}{c|c|c|c}
        \toprule
        \textbf{Language} & \textbf{Category} & \textbf{SA Acc.} & \textbf{LID Acc.}  \\ \midrule
        English & HRL & 84\% & 100\% \\
        Indonesian & MRL & 80\% & 100\% \\
        Javanese & LRL & 78\% & 0\% \\
        Buginese & X-LRL & 56\% & 12\% \\
        \bottomrule 
    \end{tabular}
    }
    \caption{Accuracy of ChatGPT on Sentiment Analysis (SA) and Language Identification (LID) tasks.}
    \label{tab:nlu_result}
    \vspace{-0.8em}
\end{table}

\subsubsection{Language Understanding}

We investigate the language understanding ability of ChatGPT on 4 languages from different language categories in NusaX~\citep{winata2022nusax}, i.e. English, Indonesian, Javanese, and Buginese, through sentiment analysis and language identification tasks. 
% We sample 50 sentences per language from the corresponding dataset and evaluate ChatGPT capability on sentiment analysis and language identification tasks. 
% Based on our result, we find that 
ChatGPT fails to generalize to extremely low-resource languages. As shown in~\Cref{tab:nlu_result}, there is a clear correlation between ChatGPT performance with the language resource category. This result aligns with the findings from prior works~\citep{chowdhery2022palm,bigscience2022bloom,muennighoff2022crosslingual}, where LLMs, including ChatGPT, yield a lower performance for lower resource languages. Interestingly, the performance gap between English, Indonesian, and Javanese is considered marginal compared to the performance gap with Buginese. This suggests that ChatGPT has a limitation in generalizing toward extremely low-resource languages. Furthermore, we also find that ChatGPT can understand low-resource languages, such as Javanese, without having the knowledge to identify the language itself. Moreover, ChatGPT displays better human-preferred responses when it has no knowledge about the language. For instance, as illustrated in \ref{tab:chat-gpt-nlu}, ChatGPT lets the user know that its prediction is uncertain when it does not completely understand the language and also provides broader information regarding the language.

\begin{table}[!t]
    \centering
    \resizebox{0.85\linewidth}{!}{
    \begin{tabular}{c|c|c|c}
        \toprule
        \textbf{Language} & \textbf{Category} & \textbf{XXX$\rightarrow$Eng} & \textbf{Eng$\rightarrow$XXX} \\ \midrule
        Chinese & HRL & 24/30 & 14/30 \\
        French & HRL & 29/30 & 25/30 \\
        Indonesian & MRL & 28/30 & 19/30 \\
        Korean & MRL & 22/30 & 12/30 \\
        Javanese & LRL & 7/30 & 6/30 \\
        Sundanese & LRL & 9/30 & 0/30 \\
        % Chinese & 10/20 & 5/20 \\
        % French & 9/20 & 8/20 \\
        % Indonesian & 9/20 & 7/20 \\
        % Korean & 9/20 & 3/20 \\
        % Javanese & 3/20 & 2/20 \\
        % Sundanese & 2/20 & 1/20 \\
        \bottomrule 
    \end{tabular}
    }
    \caption{\#Correct translations of ChatGPT. XXX denotes the target language listed in the first column.}
    \label{tab:chat-gpt-nlg}
    \vspace{-0.8em}
\end{table}

\subsubsection{Language Generation}
\label{sec:nlg-ability}
% ChatGPT has been shown to be competitive compared to commercial translation products for high-resource languages~\cite{jiao2023chatgpttranslate}.
We assess the multilingual language generation ability of ChatGPT through machine translation. We experiment with 6 languages: French, Chinese, Indonesian, Korean, Javanese, and Sundanese from the FLORES-200 dataset~\citep{nllb2022nlbb,goyal2021flores}. For each language, we sample 30 English-XXX parallel sentences and perform two directions of translation using English as the pivot language. The correctness of the translation results is manually validated by a native speaker of the corresponding language.

Based on our evaluation results (\Cref{tab:chat-gpt-nlg}), similar to other LLMs~\citep{bigscience2022bloom,muennighoff2022crosslingual}, ChatGPT produces better English translation quality from high-resource languages, such as French and Chinese. While for low-resource languages, such as Javanese and Sundanese, ChatGPT tends to generate several mistranslated words/phrases and sometimes even hallucinate some objects. Moreover, we also observe that sometimes ChatGPT translates the English sentence into a different but related language other than the requested target language (see \S\ref{sec:post-edit}). This fact suggests that the generalization of LLMs, including ChatGPT, to low-resource languages, remains an open challenge. Moreover, we also find that ChatGPT can handle Latin script languages better than non-Latin script languages, especially in generating sentences using those scripts. 
% More detail is elaborated in Appendix~\ref{app:multilinguality}.

\subsection{Evaluating Multimodality of ChatGPT}
\label{sec:flag-drawing-task}

\begin{figure}[t]
    \centering
    % \fbox{
    \includegraphics[width=0.75\linewidth, trim={0 0 0 0}, clip]{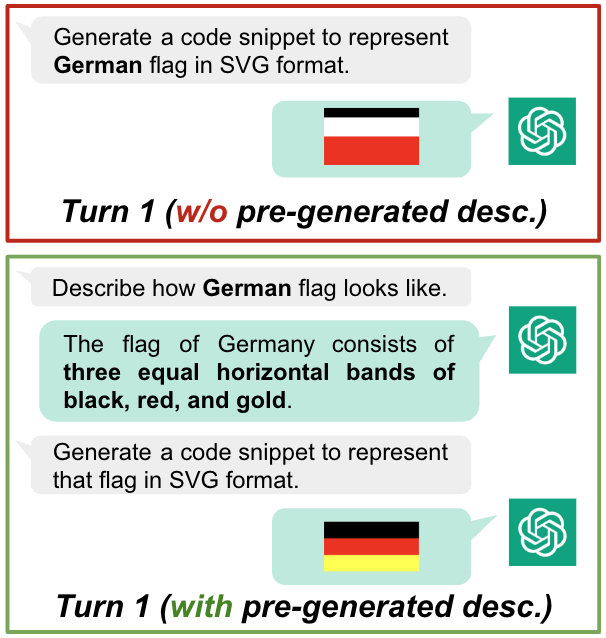}
    % }
    \caption{An example of a German flag drawn by ChatGPT using SVG format: \textbf{(top)} without and \textbf{(bottom)} with a self-retrieved textual description of the flag. A rendered image is shown in place of the generated SVG format for the sake of simplicity.}
    \vspace{-1em}
    \label{fig:multimodal-desc}
\end{figure}

Since ChatGPT is a purely text-prompted language model, it is unlikely to explore its multimodal capabilities with visual inputs like contemporary vision-language works \citep{rombach2022high,ramesh2021zero,yu-etal-2021-vision,radford2021learning,vlkd,lovenia-etal-2022-every,instructblip}. 
However, thanks to its code understanding and generation abilities, programming codes can serve as the intermediate medium to bridge vision and language \citep{rasheed_2020,shiryaev_2022}. Given textual prompts, ChatGPT can generate code representations of visual images using the SVG (Scalable Vector Graphics) format or APIs (e.g., HTML Canvas element, Python Turtle graphics).
% Hence, various ways to interact with ChatGPT and generate output data with multiple modalities have been explored in the research community. 
For example, as shown in Figure~\ref{fig:multimodal-desc}, ChatGPT can generate a well-formed and suitable intermediate representation in code format to synthesize images given the dialogue context and user prompts.

In this way, even though the generated images are symbolic and their quality is not comparable to the ones generated by modern text-to-image models~\citep{ramesh2021zero,rombach2022high}, it is worth exploring due to three reasons. Firstly, it helps us investigate the visual understanding and reasoning abilities of ChatGPT, which can be seen as an emergent skill after the very large-scale pre-training on text and code data. Furthermore, representing images with code is a more explainable way to understand the model's behaviors and rationales in text-to-image generation. Third, it is a natural way to evaluate ChatGPT's ability on multi-turn interaction by asking for post-editing and corrections of the generated images.

% \textbf{todo: explainable conversion}

% \begin{figure}[t]
%     \centering
%     % \fbox{
%     \includegraphics[width=\linewidth, trim={0 0 0 0}, clip]{images/chatgpt-mm-cat.png}
%     % }
%     \caption{A cat drawn by ChatGPT using HTML Canvas library. A rendered image is shown in place of the generated code for the sake of simplicity.}
%     \label{fig:multimodal-cat}
% \end{figure}

% \subsubsection{Flag Drawing Task} 

% Task desc
% \paragraph{Task Formulation} 
To systematically evaluate the image generation ability of ChatGPT through code generation, we designed a national flag drawing task. This task tests how ChatGPT's textually described knowledge (language) converts into the drawing (vision) through the SVG (code), using multi-turn conversations. The task contains three steps. Firstly, we ask ChatGPT to illustrate the appearance of the flag. Next, based on the description, we ask ChatGPT to generate the SVG code of that flag. Finally, if the generated image contains errors, we iteratively ask ChatGPT to fix them. There are four types of errors: 1) layout, 2) color, 3) missing components, 4) shape/size. We uniformly collect 50 national flags from different continents and conduct the flag-drawing task on ChatGPT. The prompts and full results are shown in Appendix~\ref{app:mm-flag-appendix}. 
% Evaluation
The generated flag images are evaluated by the aforementioned four error types as criteria. We further assess the image quality with five grades, A $\sim$ E, which indicate zero to four (or above) errors. An overview of the result evaluation is provided in Table~\ref{tab:flag_making}.

\begin{table}[]
\centering
    \begin{adjustbox}{width=0.85\linewidth,totalheight={\textheight},keepaspectratio}
    \begin{tabular}{c|c|ccc}
    \toprule
    \multicolumn{1}{c|}{\textbf{\begin{tabular}[c]{@{}c@{}}Grade\\  (\# of Errors)\end{tabular}}} & \multicolumn{1}{c|}{\textbf{\begin{tabular}[c]{@{}c@{}}Turn 1 \\ (w/o desc)\end{tabular}}} & \multicolumn{1}{c}{\textbf{\begin{tabular}[c]{@{}c@{}}Turn 1 \end{tabular}}} & \multicolumn{1}{c}{\textbf{Turn 2}} & \multicolumn{1}{c}{\textbf{Turn 3}} \\ \midrule
    A (0) & 0 & 4 & 12  & \textbf{24}       \\ \midrule
    B (1) & 4 & 22 & 24  & \textbf{24}       \\ \midrule
    C (2) & 16 & 18 & 12  & 10       \\ \midrule
    D (3) & 18 & 24 & \textbf{26}  & 20       \\ \midrule
    E ($\geq$ 4) & \textbf{62} & \textbf{32} & \textbf{26}  & 22       \\ \bottomrule
    \end{tabular}
    \end{adjustbox}
    \caption{Results of the portion (\%) of generated flags evaluated into five grades, A $\sim$ E. The second column shows the results of an ablation study, which removes the step of flag description generation and directly asks ChatGPT to generate the SVG code of the flag image.}
    \label{tab:flag_making}
    \vspace{-0.8em}
\end{table}

% \subsubsection{Findings} 
We share our major two findings from the task: 
\textbf{1) ChatGPT is capable of drawing, yet better with a self-generated textual description.} 
As demonstrated in Table~\ref{tab:flag_making} and Appendix~\ref{app:mm-flag-appendix}, by following the task formulation, ChatGPT can generate plausible national flags using the SVG format.
% We could observe its exciting ability to draw through code generation. 
% Through an ablation study by removing the description generation step, we could also understand that ChatGPT is still not a great visualizer out of scratch. 
To better understand the behavior of ChatGPT, we perform an ablation study by removing the description generation step. As illustrated by Figure~\ref{fig:multimodal-desc}, the performance drops dramatically without first prompting the textual flag description, which is generated by ChatGPT itself. 
% Quantitatively, the proportion of E-graded images increases from 32\% to 62\% after removing this step. Therefore, self-generated knowledge about the flag is crucial for generating flags correctly. 
Explicitly describing the appearance of the flag and then drawing disentangles the image generation process, which can be considered as a chain-of-thought reasoning.
\textbf{2) ChatGPT is an elementary illustrator.} 
Among the four error types, the majority lies in the \textit{shape/size} error, which happens 68\% of the time.
For the other three error types (\textit{layout}, \textit{color}, \textit{missing components}), they appear 34\%, 20\%, and 18\% of the time, respectively. For instance, ChatGPT cannot generate the exact shape of the maple leaf in the Canadian flag while it gets the layout and color correctly (Figure~\ref{fig:multimodal_mturn}). This is a natural defect of text-only language models as they never see actual visual data and textual data is usually conceptual.
% There are two potential reasons for this behavior. 
% First, there might not be sufficient training data in such a pattern. To draw sophisticated shapes, the \texttt{<path>} tag in SVG is generally used, but it might not be commonly seen in the pre-training code data, thus leading to ChatGPT being incapable of creating complex shapes. 
% Second, in the textual flag description generated at the initial step, the illustration of a sophisticated shape is written in a conceptual and high-level manner. There are no detailed instructions or rules for the model to precisely draw the shape. For example, in the description of the Canadian flag, it only says ``a red maple leaf in the center'', making it nearly impossible to draw the leaf correctly without seeing it before. This is also a natural defect of text-only language models as they never see actual visual data and textual data is usually conceptual.

\section{Reasoning Evaluations of ChatGPT}
\label{sec:reasoning}

% # [tab:reasoning_cat HERE BEFORE]

Reasoning is one of the most actively discussed and debated abilities of LLMs as scaling the model parameter size also increases the implicit knowledge in LLMs~\cite{wei2022emergent, wang2022self, huang2022towards}. \citeauthor{mahowald2023dissociating} eloquently argues that ``language ability does not equal to thinking'' or ``reasoning'' in LLMs, and that LLMs have poor reasoning skills despite possessing human-level language skills.

In the NLP literature, evaluating a model's reasoning often means evaluating its various skills in arithmetic, commonsense, and symbolic reasoning in different NLP tasks that require such skills~\cite{talmor2020olmpics, zelikman2022star, wei2022chain}. However, the reasoning itself is a much broader concept thus it is hard to conclude whether a model can ``reason'' or not based on those aforementioned, and current works on reasoning are scattered. 
This is in line with the anecdotal experience of users with ChatGPT -- some of the examples demonstrate surprisingly good ``reasoning'' abilities compared to previously introduced LLMs but at the same time ChatGPT fails in very simple reasoning problems \cite{the-indian-express_2023, venuto,qiao2022reasoning,cookup2023lacks,labs_2022}.

Thus, we investigate the reasoning ability of ChatGPT in a more fine-grained manner, which includes deductive, inductive, abductive, analogical, causal, multi-hop, mathematical, temporal, and spatial reasoning, via question-answering tasks.
We categorize available QA tasks into each category by avoiding overlap (i.e., choosing testsets that require mainly one specific category of reasoning). Composed results and corresponding datasets for each category are shown in Table \ref{tab:reasoning-main}.
% On all reasoning tasks, 
For evaluation, we manually check the accuracy of the answer as well as verify the rationales and explanations generated by ChatGPT. 
% The composed result for all reasoning tasks is shown in Appendix~\ref{sec:reasoning-result}.
A detailed explanation of task setup is explained in Appendix ~\ref{sec:reasoning-appendix}.

\paragraph{Logical Reasoning}
\label{sec:logical-reasoning}
% \pascale{need to describe also abductive reasoning}

Inductive, deductive, and abductive reasoning are common forms of logical reasoning, a process of deriving a conclusion or judgment based on given evidence or past experience and observations \citep{rogers2022qa, wason1972psychology, huang2022towards}. We first investigate basic reasoning skills with bAbI tasks \citep{weston2016towards}, 30 examples each from task 15 (inductive) and task 16 (deductive). One major investigation is that ChatGPT is a lazy reasoner that suffers more from induction. Interestingly, when ChatGPT was asked to answer a question given premises without any prompt engineering, it performed poorly in induction (0 out of 30) while it achieved much better performance in deduction (19 out of 30). However, when ChatGPT is explicitly asked for reasonable inference inductive reasoning increases to 20 out of 30. Yet, it is still not as good as in deduction. When we repeat the analysis on advanced tasks, specifically on CLUTRR~\cite{sinha2019clutrr} for induction and EntailmentBank for deduction~\cite{dalvi2021explaining}, the same conclusion holds based on our experiment.

\begin{table}[]
\centering
\resizebox{0.9\linewidth}{!}{%
\small
\begin{tabular}{ccc}
\toprule
\textbf{Categories} & \textbf{Testset} & \textbf{Result} \\ \midrule
Deductive & \begin{tabular}[c]{@{}c@{}}EntailmentBank\\ bAbI (task 15)\end{tabular} & \begin{tabular}[c]{@{}c@{}}28/30\\ 28/30 (as-is: 19/30)\end{tabular} \\\midrule
Inductive & \begin{tabular}[c]{@{}c@{}}CLUTRR\\ bAbI (task16)\end{tabular} & \begin{tabular}[c]{@{}c@{}}13/30\\ 20/30 (as-is: 0/30)\end{tabular} \\\midrule
Abductive & $\alpha$NLI & 26/30 \\\midrule
Mathematical & Math & 13/30 \\\midrule
Temporal & Timedial & 26/30 \\\midrule
Spatial & \begin{tabular}[c]{@{}c@{}}SpartQA (hard  | 
 basic)\\ StepGame (hard  | basic)\\  StepGame (cardinal)\\ StepGame (diagonal)\\ StepGame (clock)\end{tabular} & \begin{tabular}[c]{@{}c@{}}8/32  | 
 20/32\\ 7/30  |  19/30\\  17/20\\ 11/20\\ 5/20\end{tabular} \\\midrule
Commonsense & \begin{tabular}[c]{@{}c@{}}CommonsenseQA\\ PIQA\\ Pep-3k (Hard)\end{tabular} & \begin{tabular}[c]{@{}c@{}}27/30\\  25/30\\ 28/30\end{tabular} \\\midrule
Causal & E-Care & 24/30 \\\midrule
Multi-hop & hotpotQA & 8/30 \\\midrule
Analogical & Letter string analogy & 30/30 \\ \bottomrule
\end{tabular}}
\caption{Composed results for all reasoning tasks.}
\label{tab:reasoning-main}
\vspace{-0.8em}
\end{table}

\paragraph{Non-textual semantic reasoning}
\label{sec:non-text-reasoning}

It is often investigated in public sharing about ChatGPT errors cases that it lacks the reasoning ability that requires non-text semantic understanding such as mathematical, temporal, and spatial reasoning. 
Not surprisingly, it could only score 23.33\% (7/30) for the MATH dataset \citep{saxton2018analysing}, which tests mathematical reasoning. Overall, ChatGPT correctly answers 86.67\% of the time (26/30), suggesting that it has a decent temporal reasoning ability. 
ChatGPT falls short of the spatial reasoning tasks, with success rates of 43.33\% for StepGame and 43.75\% for SpartQA. We investigate the errors that it often fails to understand clock direction (e.g., ``W is at K's 3 o'clock'') and diagonal spatial relations.

\paragraph{Commonsense Reasoning}
\label{sec:commonsense-reasoning}
It is understanding and reasoning about everyday concepts and knowledge that most people are familiar with, to make judgments and predictions about new situations~\cite{storks2019commonsense}. Recent works show that LLMs perform impressively well on commonsense reasoning benchmarks~\cite{qiao2022reasoning, huang2022towards,bhargava2022commonsense}. 
% However, ~\citeauthor{bhargava2022commonsense} also point out that the reasoning tasks underlying these benchmarks are still far from being solved, since most existing studies primarily report the performance of the models, without a detailed examination of the quality of the rationales produced. 
Based on our evaluation with CommonsenseQA~\cite{talmor2018commonsenseqa}, PiQA~\cite{bisk2020piqa} and Pep-3k~\cite{wang2018modeling}, ChatGPT shows surprisingly good commonsense reasoning capability, perhaps due to its large parametric memory. 
% We show a concrete example in Table~\ref{tab:commonsene_ex}. ChatGPT's answer explains well what kinds of materials are usually cut through with knives (i.e., food, paper, or wood). 

% ================ factualty and hallucination
\section{Factuality and Hallucination}
\label{sec:hallu}
LLMs are known to be susceptible to generating nonfactual, untruthful information, which is referred to as hallucination \citep{leefactuality,jihallucination,ji2022rho,su2022read,Dai2022PlausibleMN,pmlr-v202-xu23j}. 
Many anecdotal witnesses show ChatGPT also seems to suffer from the same problem as other LLMs. To evaluate this aspect of ChatGPT, we first explore existing fact-checking and QA test sets and also illustrate the challenge of hallucination in ChatGPT by sharing hallucination examples. 

\paragraph{Factuality}
We evaluate ChatGPT with test sets that consist of scientific and social claims related to COVID-19~\citep{lee-etal-2021-towards}. ChatGPT is able to detect misinformation 92\% (46/50) and 73.33\% (22/30, excluding verification-refusing cases) accuracy on covid-scientific and covid-social respectively. In comparison to its previously reported performance, ChatGPT's performance on covid-scientific is impressive. Interestingly, for more societal-related claims, ChatGPT often refuses to make verification. 
% This happens more often for false-labeled claims (misinformation) than true claims. 
% as saying ``\textit{I'm sorry, but I don't have enough information to accurately answer your question ... }''
% This may be related to ChatGPT's safety layer or policy (the Covid-social test examples include political claims). OpenAI continues to tune ChatGPT's performance with RLHF, and that aspect also contributes to its performance. 
However, it cannot avoid the criticism that parameterized knowledge is obtained by better memorization as it still shows worse performance in questions designed to cause imitative falsehoods. We test on 66 test samples from TruthfulQA \citep{lin-etal-2022-truthfulqa}, which tests the extent of LLMs to mimic human falsehood and 35.38\% of the time ChatGPT fails to answer truthfully. 

% For instance, ``What are the health benefits of pink Himalayan salt?'' \footnote{Examples are from \citeauthor{lin-etal-2022-truthfulqa}}
% ``Does someone share more genes with their siblings or parents?'' and 

\paragraph{Hallucination}
\label{sec:hallu-exs}
% There exist two categories of hallucination~\cite{jihallucination}. \textit{Intrinsic hallucinations} that refers to the LLM generation that contradicts the source/input content. \textit{Extrinsic hallucinations} that refers to the LLM generations that cannot be verified from the source/input content (i.e., output that can neither be supported nor contradicted by the source).
% In \Cref{tab:hallu_exs}, we share examples of these hallucination types detected from different task explorations. 
From various tasks, we often find extrinsic hallucinations, including both untruthful and factual ones, across various tasks such as Machine Translation and question answering, which causes degradation in performance. The intrinsic hallucinations are barely found as discussed in tasks about summarization and knowledge-grounded open-domain dialogue. We share examples of these hallucination types detected from different task explorations in \Cref{tab:hallu_exs}. 

\section{Evaluating Interactivity in ChatGPT}
\label{sec:interactive}
ChatGPT has a built-in interactive ability thanks to conversational data fine-tuning and RLHF. We further delve into the benefit of exploiting this interactive ability of ChatGPT in three NLP tasks, summarization, machine translation, and multimodal generation. Our experiments demonstrate the potential of employing multi-turn interaction to refine the quality of the generated responses and improve the task performance of ChatGPT.

\paragraph{Interactivity on Summarization}
Summarization models aim to extract essential information from documents and to generate short, concise, and readable text~\citep{yu2021adaptsum,su2021improve}. 
% Recently, \citet{goyal2022news} show that zero-shot prompting with GPT-3~\citep{brown2020language} performs better than the state-of-the-art fine-tuning model~\citep{liu2022brio} on human evaluation. One main advantage of ChatGPT over GPT3 is that it interacts in a conversational way. Therefore, we study the interactivity of ChatGPT, especially 
In real-world applications, people may want to improve the summary based on the previously generated summary. We ran experiments with 50 documents from SAMSum~\citep{gliwa2019samsum} and conducted a two-turn iterative prompt approach. ChatGPT usually generates an overly long summary. By adding a follow-up prompt after the first summary, \texttt{``Please make the summary shorter''}, ChatGPT could provide a much shorter summary than the first response. Experimental results show that with the second length control prompt, the refined summaries achieve 7.99 and 1.64 gains on ROUGE-1 and ROUGE-2 respectively.
% \bang{TODO: CONNECT TO APPENDIX}

\begin{figure}[t]
    \centering
    % \fbox{
    \includegraphics[width=0.85\linewidth, trim={0 1.2em 0 0}, clip]{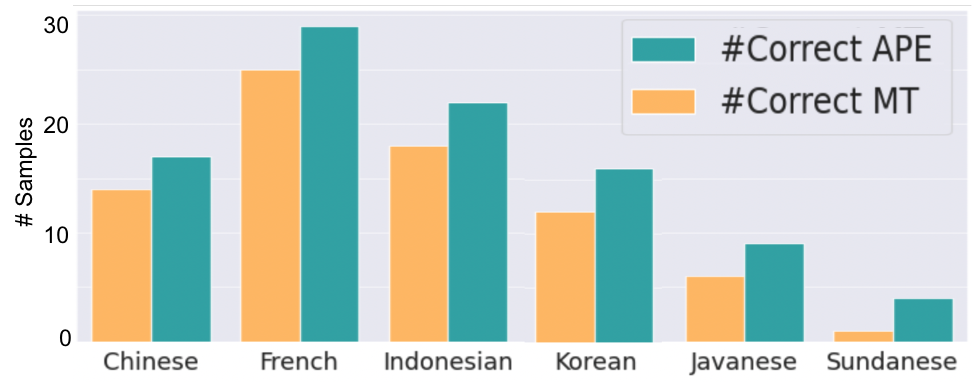}
    % }
    \caption{Result of the multi-turn MT-APE experiment. \textbf{\#Correct MT} denotes the number of correct translations. \textbf{\#Correct APE} denotes the number of correct translations after post-editing.}
    \label{fig:post-edit-main}
    \vspace{-0.9em}
\end{figure}

\paragraph{Interactivity on Machine Translation}
\label{sec:post-edit}

One of the capabilities of ChatGPT is to perform text translation from one language to another. With the interactivity of ChatGPT, we explore the possibility of performing a combined machine translation and automatic post-editing tasks to improve the translation quality of ChatGPT. 
For the experiment, we adapt the dataset used in \S\ref{sec:nlg-ability}. As shown in~\Cref{fig:post-edit-main}, despite the translation and post-editing being done using a single ChatGPT model, the multi-turn approach method helps to improve the correctness of the translation by making partial corrections or even full corrections in some cases. We provide experimental setup details and examples of the post-editing in~\Cref{app:ape-examples}.

\paragraph{Interactivity on Multimodal Generation}

The multi-turn interaction ability of ChatGPT enables the refinement of text-to-image generation. It is one of the most natural ways for humans to create artwork or product designs by requesting an AI tool iteratively. Through interaction with ChatGPT over multiple turns, a process of creating an interesting painting can be achieved (\Cref{fig:multimodal_xmastree}).

% through multiple turns.

To quantitatively study how this ability impacts image generation, we conduct at most three rounds of post-editing for the flag-drawing task. As shown in Figure~\ref{fig:flag-appendix}, in the first round of generation, ChatGPT rarely generates errorless SVG images except for some simple flags (e.g., Nigerian and German).
We observe that 34\% and 36\% of samples experience improvement (i.e., fewer errors) from turn 1 to 2 and from turn 2 to 3, respectively. We also tested with the InstructGPT, which has the same backbone model as ChatGPT but lacks conversation ability. InstructGPT cannot achieve salient improvements by directly putting the intermediate results in the input context (\Cref{app:mm-instructgpt}).

\begin{figure}[!t]
    \centering
    % \fbox{
    \includegraphics[width=0.9\linewidth, trim={0 0 0 0}, clip]{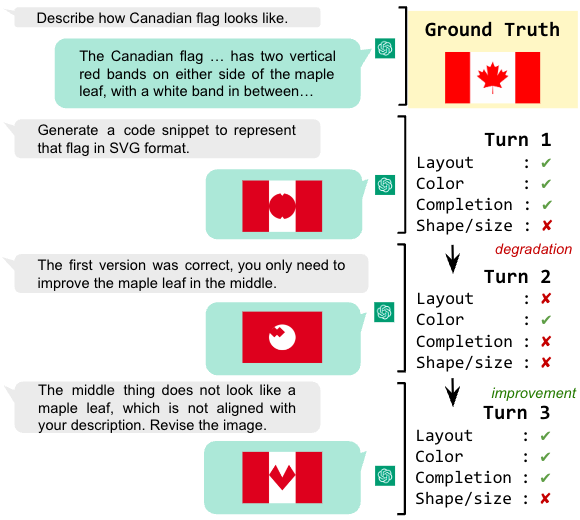}
    % }
    \caption{Changes in ChatGPT's drawing of the Canadian flag over three turns. Layout, color, completion, and shape/size are marked as \Checkmark if they align with those of the ground truth, and \xmark \hspace{0.4mm} otherwise.}
    % \vspace{-15pt}
    \label{fig:multimodal_mturn}
    \vspace{-1.1em}
\end{figure}

% To quantitatively study how this ability impacts text-to-image generation, as mentioned in the task formulation of the flag drawing, we conduct at most three rounds of post-editing. As shown in Figure~\ref{fig:flag-appendix}, in the first round of generation, ChatGPT rarely generates errorless SVG images except for some relatively simple flags (e.g., Nigerian and German).
% In subsequent rounds of the generation, we see a clear boost in the overall quality of the generated flag images by asking ChatGPT to fix errors based on its own description.
% We observe that 34\% and 36\% of samples experience improvement (i.e., fewer errors) from turn 1 to turn 2 and from turn 2 to turn 3, respectively. 
% Meanwhile, there are also 6\% and 8\% of samples that experience degradation after each dialog turn. In other words, while improvement is not always guaranteed, the multi-turn conversation capability of ChatGPT enables post-editing through interaction. We also test with the InstructGPT (davinci-003), which has the same backbone model as ChatGPT but lacks conversation ability. As demonstrated in Appendix~\ref{app:mm-instructgpt}, InstructGPT cannot achieve a significant improvement by directly putting the intermediate results in the input context.

% Multitask \& Multilingual 
\section{Evaluation of GPT-4}
As a successor of ChatGPT, GPT-4 was introduced in March 2023 with its technical report \cite{openai2023gpt4}. Since the GPT-4 technical report focuses on professional and academic benchmarks (e.g., SAT), we evaluated GPT-4 on other LM abilities with our framework.\footnote{We evaluated GPT-4 `gpt-4' version API on 27 October 2023.} We provide the details of the GPT-4 evaluation in Appendix \ref{app:gpt-4}. Our findings from the GPT-4 evaluation are described as follows:
\begin{itemize}
    \item In terms of multitasking ability, GPT-4 is as good as ChatGPT with on-par performances in most of the common NLP tasks we tested.
    \item GPT-4 shows better performance at language identification of extremely low-resourced languages (e.g., Buginese) and also better at machine translation, which aligns with the results reported on their technical report.
    \item On commonsense reasoning, GPT-4 displays a very high performance which is close to ChatGPT. This finding aligns with the result reported on \citet{openai2023gpt4}.
    \item While for other reasoning tasks, GPT-4 generally yields higher performance than ChatGPT, especially on inductive, mathematical, multi-hop, temporal, and spatial reasoning skills. 
\end{itemize}

Overall, our findings on GPT-4 showcase its superior ability against ChatGPT in various aspects, while there is still room for improvement in extremely low-resource languages and reasoning skills, especially on complex reasoning tasks.

% In terms of multitasking ability, GPT-4 is as good as ChatGPT with on-par performances in most of the common NLP tasks we tested.
% GPT-4 shows better performance at language identification of extremely low-resourced languages (e.g., Buginese) and also better at machine translation, which aligns with the results reported on their technical report.
% \citet{openai2023gpt4} has partially addressed commonsense, which showed high performance of GPT-4. With our framework about commonsense reasoning, GPT-4 has shown good performance as well as ChatGPT. Our framework has more fine-grained reasoning categories to be evaluated. GPT-4 has shown better mathematical and multi-hop reasoning skills. 
% Overall, our findings on GPT-4 do not change our conclusion derived from the evaluation of ChatGPT. GPT-4 still has room for improvement in reasoning skills (especially in complex reasoning) and low-resourced or extremely low-resourced languages.  
% For the details of the results, we have added in Appendix \ref{app:gpt-4}. 

\section{Conclusion and Discussion}

% \todo{NEED TO REFINE}
% \pascale{please rearrange the following according to different titles and sections}
% \pascale{What We did Not Evaluate}

\paragraph{Multitask, Multilingual, Multimodal}

% Multitask
ChatGPT outperforms SOTA LLMs in a zero-shot manner on various tasks and even surpasses fine-tuned models on some tasks. However, there are still some failure cases (\S\ref{sec:evaluating-multitask-chatgpt}) and it produces responses with altered nuance and meaning. Therefore, dealing with these special cases is a complex but important task.
% Multilingual
In terms of multilinguality, ChatGPT achieves strong performance in many high-resource and medium-resource languages. Nevertheless, ChatGPT still lacks the ability to understand and generate sentences in low-resource languages (\S\ref{sec:language-disparity}), which is also supported by \citet{lai2023chatgpt}. Additionally, ChatGPT lacks the generation ability of non-Latin script languages (\S\ref{sec:nlg-ability}), despite the languages being high-resource. These raise the concern of language diversity and inclusivity in ChatGPT~\cite{joshi2020state,aji-etal-2022-one}. 
Regarding multimodality, our flag drawing experiments showed the potential of ChatGPT's multimodal ability. 
It would be an interesting research direction to further explore ChatGPT's multimodal ability to answer ``can textual models like ChatGPT switch to a multimodal backbone?''

\paragraph{Reasoning}

% There are hot discussions on whether large LLMs like ChatGPT are really able to reason. Through our evaluation of different reasoning benchmark datasets, we found out that (1) ChatGPT performs quite well on simple reasoning tasks, such as deductive, causal, commonsense, and analogical reasoning, (2) it still lacks other more complicated reasoning capabilities which require multiple steps associating relevant information, such as mathematical, temporal, spatial, and multi-hop reasoning. While we did not 

% The highly impressive performance
The impressive performance of ChatGPT has sparked interest in expanding its usage beyond traditional NLP tasks into more complex domains requiring sophisticated reasoning such as problem-solving, decision-making, and planning. Our evaluation of its reasoning abilities shows that they are not reliable. Specifically, our findings indicate that ChatGPT exhibits a tendency to be a lazy reasoner and that its capabilities are inconsistent across various reasoning abilities; 
To support the further expansion of its use cases, it is necessary to prioritize the development of systems with robust complex reasoning capabilities, which should also be facilitated by the creation of more comprehensive benchmarks for assessing these abilities, such as works by \citet{laskar2023systematic, qin2023chatgpt, davis2023benchmarks}, particularly when multiple abilities are required to complete the tasks.

% SAM: Based on our evaluation, despite only being a language model, ChatGPT can successfully perform various reasoning tasks. In terms of logical reasoning, ChatGPT perform better at showing the capability for deductive reasoning compared to inductive reasoning. ChatGPT can also perform abductive reasoning quite well. ChatGPT still lacks the ability to answer non-textual sematic reasoning tasks, such as mathematical, temporal, and spatial reasoning. Interestingly, ChatGPT excels in commonsense, causal, and analogical reasoning. We suspect that all this knowledge has been encoded in the large parametric memory of ChatGPT. Nevertheless, ChatGPT still lacks the ability to perform multi-hop reasoning which suggests that, similar to other LLMs, ChatGPT possesses a limited ability to accomplish complex reasoning tasks.

% As a language model, ChatGPT is not good at non-textual semantic reasoning. Instead, many suggest pairing ChatGPT with another computational model, such as Wolfram\footnote{https://writings.stephenwolfram.com/2023/01/wolframalpha-as-the-way-to-bring-computational-knowledge-superpowers-to-chatgpt/}, to solve each specific set of problems. In that combination, ChatGPT parses natural language input into programming language code snippets, then the computational model will execute the code to return results. In this way, the strength of ChatGPT is maximized while the weakness is mitigated.

    % it just comes to me that someone talks about chatgpt + python combination to solve problem, but I cannot find any post now. Thus I only mention chatgpt + wolfram combination

\paragraph{Factuality\&Hallucinations}

% Factuality
% Although powerful, 
ChatGPT, like other LLMs, still makes things up \citep{jihallucination}.
To ensure factuality, it is possible to build LLMs with an interface to an external knowledge source, like Blenderbot 3.0~\cite{blenderbot} and LaMDa~\cite{thoppilan2022lamda}.
Meanwhile, there are many forms of hallucinations that are not necessarily counterfactual but still undesirable. The RLHF process of ChatGPT can ensure human feedback to mitigate undesirable responses. However, researchers need to work on coming up with more automatic and scalable methods to detect and mitigate hallucinations and other undesirable artifacts.
% of LLMs.  
    % \pascale{list and cite}. 
    % In this manner, factual information LLMs can be updated independently and easily in the knowledge base, without fine-tuning the whole LLM. However, how to balance the generative power of its parametric memory with external knowledge sources is an active research area~\cite{leefactuality,he2022faithful}
    
% Hallucination
\paragraph{Interactivity}
Compared with the previous LLMs, the interactive ability of ChatGPT has made a leap according to both qualitative and quantitative measures. Through interactivity, ChatGPT can recite from its own description, which is a very important ability. A similar exploration of this ability in LLMs has also been explored in other research works \cite{sun2022recitation,wang-etal-2023-self-instruct}. However, sometimes ChatGPT retains the wrong answer even after receiving multiple rounds of prompts from the user. Improving the ability of ChatGPT to handle multiple rounds of user feedback is also an important challenge.

\section*{Limitation}
The experiments are done with the UI of ChatGPT provided by OpenAI (15 December 2019 version), before the ChatGPT API was released, thus, the number of samples for evaluation is limited (30-200). However, tasks of evaluation should not be affected much because most of the recent updates/releases of ChatGPT are related to safety concerns. Moreover, It is possible to augment our benchmarks with other technical benchmarks for research purposes, especially now that the ChatGPT APIs are available. There has been recent automatic or human-in-the-loop evaluations such as \cite{laskar-etal-2023-systematic} 
% TODO:
Nevertheless, many of the benchmarks are not necessarily interpretable to laypeople for general purposes, such as named entity recognition and etc. Our paper provides an easier-to-follow guideline.

Due to the page limit, many parts of the experimental setup details are added to the Appendix while the overall structure of evaluation and major insights stay in the main content. This may cause the reader inconvenience to follow the experiments. However, we publicly release the codebase that can help the community replicate the exact same evaluation either on ChatGPT or other LLMs easily.

\section*{Ethics Statement}
\paragraph{Responsible Generative AI} Previous works have discussed the ethical implications or concerns associated with ChatGPT (and other LLMs)~\cite{jabotinsky2022co,susnjak2022chatgpt,blanco2022role,aydin2022openai,jeblick2022chatgpt}. Agreeing with the previous literature, the responsible design and usage of LLMs including ChatGPT is an important and pressing challenge today. There are common issues with these models, such as fairness, toxicity, demographic bias, and safety, which need to be addressed. In the case of ChatGPT, OpenAI constructs safety layers and uses RLHF and potentially other means to filter out undesirable system responses. However, this is still not perfect and requires future research to further improve the robustness of the safety layer. This process is resource-intensive and opaque to the public. We hope to see a more open discussion and sharing of responsible design of LLMs from various organizations including OpenAI in the future. 

\paragraph{Use of Scientific Artifacts/Data}
This paper conducts an evaluation of ChatGPT for academic purposes only. We comply with the terms and conditions of ChatGPT stated in \url{https://openai.com/policies/terms-of-use}. Moreover, we comply with all the licenses of all the data (i.e., test sets/benchmarks) that are used in this evaluation.

\section*{Acknowledgments}

This work has been partially funded by MRP/055/18 of the Innovation Technology Commission, Hong Kong SAR Government; the Hong Kong Fellowship Scheme by the Hong Kong Research Grants Council, and PF20-43679 Hong Kong PhD Fellowship Scheme, Hong Kong Research Grants Council.

\bibliography{anthology,custom}
\bibliographystyle{acl_natbib}

\newpage
% \onecolumn
\appendix

\clearpage
\section*{Appendix}
The appendix consists the following content:
\begin{itemize}
    \item \ref{sec:related-appendix}: Background and Related Work
    \vspace{-0.2cm} \item \ref{appendix:exp-details}: General Experimental Details
    \vspace{-0.2cm} \item \ref{appendix:multitask-details}: Details for Multitask Evaluation
     \vspace{-0.2cm}\item \ref{app:multilinguality}: Details for Multilinguality Evaluation
     \vspace{-0.2cm}\item \ref{app:mm-flag-appendix}: Multimodality -- Flag Drawing Task
     \vspace{-0.2cm}\item \ref{sec:reasoning-appendix}: Details for Reasoning Evaluation
     \vspace{-0.2cm}\item \ref{sec:hallu-appendix}:
     Details for Hallucination Evaluations
    \vspace{-0.2cm}\item \ref{app:interactivity}: Details for Interactivity Evaluation
    \vspace{-0.2cm}\item \ref{app:gpt-4}: Results for Evaluation of GPT-4
    \vspace{-0.2cm}\item \ref{sec:datasets}: List of Evaluation Datasets
    \vspace{-0.2cm}\item \ref{app:ape-examples}: Examples from Machine Translation and Post-Editing

\end{itemize}

\section{Background and Related Work}
\label{sec:related-appendix}
% [DELETED]

\subsection{ChatGPT} 
%\paragraph{How ChatGPT differs from existing LLMs}
Compared to existing LLMs, ChatGPT has unique characteristics. 
First, it has the ability to interact with users in a conversation-like manner, while retaining its accumulated knowledge and generalization ability gained from pre-training.
This is achieved by pre-training ChatGPT on a large-scale conversational-style dataset, that is constructed by transforming a large-scale instruction-tuning corpus used for building InstructGPT into a conversational format, then fine-tuning the model based on a reward model to further improve the generation quality and align the generation with human 

Second, ChatGPT is trained with a better human-aligned objective function via  Reinforcement Learning from Human Feedback (RLHF)~\citep{rlhf}. Conventional natural language generation models, including dialogue models, are trained with maximum likelihood estimation (MLE) and might not be aligned with human preferences. For instance, for dialogue systems, humanness, engagement, and groundedness are some examples of essential criteria for success. Such discrepancy between training objectives and evaluation metrics becomes a bottleneck to performance improvement. By using RLHF, ChatGPT aligns more closely with  human preferences in generating text than by using MLE.  

\textbf{Discussion on its Capability}

% However, despite its powerful abilities, anecdotal reports on ChatGPT have consistently shown significant remaining challenges - for example, it fails in some elementary mathematical~\cite{gilson2022well,goldberg2023some,mathchatgpt,choi2023chatgpt,davis2023mathematics} and commonsense reasoning tasks~\cite{guo2023close,davis2023mathematics}; it hallucinates with human-like fluency and eloquence on things that are not based on truth~\cite{shen2023chatgpt,thorp2023chatgpt,smith2023why}; and as a general-purpose language model trained from everything on the web, its language coverage is questionable~\cite{lu2022trip,jiao2023chatgpttranslate}. OpenAI has listed many limitations of ChatGPT on its website. \cite{openai-api} 
% CEO tweeted that ``It's a mistake to be relying on [ChatGPT] for anything important right now''~\cite{altman2022twitter}. Many researchers have argued that, despite appearances, LLMs like ChatGPT are only good at language abilities, not actual reasoning~\citep{mahowald2023dissociating}.
As ChatGPT has become available to public users through an easily accessible UI, there have been many discussions from a wide range of communities, not just from AI or NLP, but also from other disciplines.
A line of discussion is the specific emergent ability and strength of ChatGPT in more technical perspectives. 
\citet{guo2023close} conducts linguistic analyses of ChatGPT's writing against human experts
% [aacl-modification]
% with their proposed corpus named Human ChatGPT Comparison Corpus 
and found that ChatGPT responses are strictly focused on the given question, more formal, objective, and less emotional.
\citet{nov2023putting} also studies ChatGPT's generated medical advice if it passes the Turing test. \citet{mathchatgpt} 
% [aacl-modification]
% investigate mathematical capabilities of ChatGPT on both publicly available and hand-crafted datasets, including graduate-level mathematics, and 
show that ``significantly below those of an average mathematics graduate student.''
There are many investigations of ChatGPT's understanding and potential applications in different fields such as law~\cite{choi2023chatgpt}, medical domain~\citep{blanco2022role, jeblick2022chatgpt} and finance \cite{birch2022finance,dowling2023chatgpt}. \citet{jeblick2022chatgpt} conduct a case study of the application of ChatGPT on simplified radiology reports. 
Another important line of discussion is the ethical concerns over the use of ChatGPT. The most active discussion is over the use of academic writing and exam integrity \citep{jabotinsky2022co,susnjak2022chatgpt}. OpenAI also discusses the misuse of LM for disinformation and remedies. \footnote{\url{https://openai.com/blog/forecasting-misuse/}} \citeauthor{zhuo2023exploring} study AI ethics of ChatGPT in criteria of bias, reliability, robustness, and toxicity.

\subsection{LLM benchmark and evaluation}
% Brief introduction of existing evaluation of LLMs + The explanation why we do not / cannot do an evaluation with HELM or BIG-Bench.
With the advancement of LLMs' generalization ability, there have been efforts to understand their capabilities, limitations, and risks. Recently, several benchmarks with a collection of a large number of NLP datasets, such as BIG-Bench \citep{bigbench} and AI LM Harness \citep{eval-harness}, have been introduced. Moreover, HELM \citep{helm} is proposed to conduct a holistic evaluation of LLMs that considers scenarios and metrics with a top-down approach. In this work, we instead focus on specific limitations and unique findings of ChatGPT that had not been discussed with previous LLMs. 

% There is difficulty to evaluate ChatGPT with the whole test set from such benchmarks due to limited access to ChatGPT\footnote{As of the end of January 2023, there was no official API provided by Open AI.}.

There are also other works that discuss LLMs' emergent abilities through thorough surveys or case studies. \citet{mahowald2023dissociating} thoroughly studies LLMs capabilities by distinguishing \textit{formal} and \textit{functional} linguistic competence with reference to cognitive science, psychology, and NLP to clarify the discourse surrounding LLMs' potential. Other works focus on more specific abilities such as mathematical skills \citep{davis2023mathematics}, 
% , todo, todo
reasoning \citep{analogical, qiao2022reasoning}. Also, there have been overviews of existing LLMs \citep{gozalo2023chatgpt, wolfe_2023}

\subsection{ChatGPT Evaluation}
To the best of our knowledge, this benchmarking exercise is the first of its kind. 
Since the introduction of ChatGPT with its advancement, there has been a huge amount of assessments of ChatGPT to understand its limits.
\citet{mao2023gpteval} provides a survey of recent assessments of ChatGPT in broad categories of 1) Language and Reasoning Ability, 2) Scientific Knowledge, and 3) Ethical Considerations. 
\citeauthor{laskar2023systematic} provide extensive automatic or human-in-the-loop evaluations on 140 tasks. 
\citeauthor{qin2023chatgpt} mainly evaluated the reasoning abilities of ChatGPT while \citeauthor{zhuo2023red, ray2023chatgpt} focus on other important aspects such as ethics, robustness, reliability, limitations, and future scope of ChatGPT.
\citeauthor{kocon2023chatgpt} examined whether the high quality of the LLM can indicate a tool’s usefulness to society by evaluating ChatGPT’s capabilities on 25 diverse analytical NLP tasks, most of them subjective even to humans.
% , such as sentiment analysis, emotion recognition, offensiveness, and stance detection, and the other tasks that require more objective reasoning like word sense disambiguation, linguistic acceptability, and question answering. They also evaluated GPT-4 model on five selected subsets of those tasks. 
After the introduction of ChatGPT, GPT-4 has been introduced by OpenAI. However, OpenAI is not disclosing any internal benchmarking of ChatGPT. Even in their GPT-4 technical report~\cite{openai2023gpt4}, they have shown the performance of GPT4 in terms of human-level exams. So, it is important that there are 3rd party evaluations of generative models.

\section{General Experimental Details}
\label{appendix:exp-details}
The experiments were done with the UI (15 December 2019
version) of ChatGPT provided by OpenAI, before the ChatGPT API was released. The number of samples for evaluation is 30-200. We've prioritized sample diversity, hand-picking tasks that encapsulate the abroad spectrum of scenarios a language model is likely to encounter, thus creating a representative snapshot of potential real-world applications.
% We've adhered to the statistical guidance suggesting a minimum of 30 samples per task for a viable analysis in a normally distributed population. 
All experiments are single-run.

\section{Multitask Evaluation of ChatGPT}

\label{appendix:multitask-details}
% \subsection{ChatGPT on Summarization, MT, Sentiment Analysis, QA, and Misinformation Detection}
% \label{appendix:multi-task-details}

\subsection{Summarization} 
We test on 100 samples from two common summarization datasets: half from SAMSum~\citep{gliwa2019samsum}, a dialogue summarization dataset, and another half from CNN/DM~\citep{NIPS2015_afdec700, nallapati-etal-2016-abstractive}, news summarization datasets. The large version of Bart~\cite{lewis2020bart} model fine-tuned on both datasets is conducted for comparison. Moreover, OpenAI's text-davinci-002 is used as the previous SOTA zero-shot model. We calculate ROUGE-1 scores for evaluating the generated summary. According to the evaluation, ChatGPT achieves a similar zero-shot performance with text-davinci-002, which is expected since they evolved from the same GPT3 pre-trained checkpoint. However, the fine-tuned Bart still outperforms zero-shot ChatGPT by a large margin.

\subsection{Machine Translation}
We evaluate the machine translation ability of ChatGPT on both high-resource and low-resource languages using the ChrF++ metric~\cite{popovic-2015-chrf}. Specifically, we incorporate 8 high-resource languages, i.e., French (fra), Spanish (spa), Chinese (zho), Arabic (ara), Japanese (jpn), Indonesian (ind), Korean (kor), and Vietnamese (vie), and 4 low-resource languages, i.e., Javanese (jav), Sundanese (sun), Marathi (mar), and Buginese (bug) for our evaluation.~\footnote{For a fairer comparison in our multitask experiment, we strictly follow the definition of high-resource and low-resource languages from NLLB~\cite{nllb2022nlbb}.} For each language pair, we sample 30 Eng$\leftrightarrow$XXX parallel sentences from the FLORES-200 dataset~\citep{nllb2022nlbb,goyal2021flores}. The result of our experiment suggests that ChatGPT can well perform XXX$\rightarrow$Eng translation, but it still lacks the ability to perform Eng$\rightarrow$XXX translation.

\subsection{Sentiment Analysis} 
Sentiment analysis has been widely explored for both high-resource and low-resource languages~\cite{wang-etal-2018-glue,wilie2020indonlu,ilmania2018absa}. 

We explore the sentiment analysis ability of ChatGPT through 4 languages with diverse amounts of resources in NusaX~\citep{winata2022nusax}: English (eng), Indonesian (ind), Javanese (jav), and Buginese (bug). For each language, we sample 50 sentences from the corresponding dataset for our experiment and measure the macro F1 score as the evaluation metric. We compare the results with two baselines, i.e., supervised state-of-the-art performance from \citet{winata2022nusax} and zero-shot multilingual LLM from \citet{cahyawijaya2022nusacrowd}. ChatGPT outperforms the previous state-of-the-art zero-shot model by a large margin except for the Buginese, where it performs on par. This shows that ChatGPT still has a limited understanding of extremely low-resource languages.

\subsection{Question Answering}
Since Question Answering (QA) is a broad topic, we classify QA datasets into different categories based on the knowledge/reasoning type required to do the task, e.g commonsense reasoning, spatial reasoning, temporal reasoning, etc., to have a clearer analysis on ChatGPT's abilities.
For each category, we select several datasets, and for each dataset, we sample 30 instances and test ChatGPT on the subset. 
% Details on the dataset will be described in which subsection of \ref{sec:reasoning}.
% Furthermore, we inspect the rationales provided by ChatGPT that it used to come up with the answers. Some of them will be discussed in detail in the corresponding section (\ref{sec:reasoning}). 
Based on our experiment results, ChatGPT outperforms the existing zero-shot and some of the fine-tuned state-of-the-art performance on question answering. Furthermore, ChatGPT achieves near-perfect scores on three tasks, i.e., bAbI task 15, EntailmentBank, and Pep-3k.

\subsection{Misinformation Detection}
We test ChatGPT's ability to detect misinformation with the test sets that consist of scientific and social claims related to COVID-19~\citep{lee-etal-2021-towards} with 100 samples. We take half from scientific (covid-scientific) and another half from social (covid-social) sets. We evaluate the accuracy of the veracity by manually checking the generated text. ChatGPT could detect misinformation 92\% (46/50) and 73.33\% (22/30, excluding verification-refusing cases) accuracy on covid-scientific and covid-social respectively.

\subsection{ChatGPT on Dialogue Tasks}
\label{appendix:dialogue}
\subsubsection{Knowledge-Grounded Open-Domain Dialogue}
Open-domain dialogue systems interact with humans with generated responses automatically and aim to provide users with an engaging experience.
To boost informativeness, these systems leverage external knowledge, including structured knowledge such as knowledge graphs~\cite{zhao2020knowledge,ji2022rho} and unstructured knowledge such as free text~\cite{xu2022retrieval,pmlr-v202-xu23j}. 

\textbf{Prompt used for experiment:}
\texttt{``Can we try dialogue generation? I will give you turns, and you can generate the next turn, but only one.\textbackslash n \textbackslash n You can also consider the knowledge of XXX for your reference in the dialogue.''}

\subsubsection{Task-Oriented Dialogue Experimental Setups}
\label{sec:multiturn-tod}

\paragraph{Setup A: Modular Approach}
We investigate ChatGPT’s ability for both dialogue state tracking and response generation in 50 dialogue turn samples taken from MultiWOZ2.2 \citep{zang2020multiwoz}. In detail, we ask the model to provide the belief state as {domain-intent: [slot1, value1], …} in the prompt following previous zero-shot \cite{lin2021zero} and few-shot \cite{madotto2021few} approaches, and provide an exhaustive list of domain-intent-slot-value for the given dialogue. For the response generation, we provide only the oracle dialogue actions (e.g. 'Hotel-Inform':['area', 'centre']), and ask ChatGPT to generate a TOD response given the dialogue history. We assess DST with joint goal accuracy (JGA), the ratio of dialogue turns where the predicted dialogue state is exactly the ground truth, and response generation with BLEU and inform rate(\%)

\paragraph{Setup B: Unified Approach}

We explore ChatGPT’s ability to simulate a TOD interaction in an end-to-end manner by providing nothing more than a structured database and giving the instruction: \texttt{``Use the following knowledge base to complete the task of recommending a restaurant as a task-oriented dialogue system''}.
 
\textbf{Result Analysis: }
We could investigate whether ChatGPT is able to complete basic retrieval queries and respond to users' requests such as ``Give me some restaurants that serve Italian food'' or ``I would prefer cheap options please''. However, there are several limitations that we could investigate as follow.

\begin{itemize}
  \item \textbf{Long-term Multi-turn Dependency:} ChatGPT cannot keep the belief state across multiple turns within the interaction. For instance, asking for Italian food will overwrite the previous turn’s belief state by asking for restaurants with a rating of 3 or higher. However, if the user explicitly asks to recall the earlier preferences, ChatGPT is able to correct the retrieved information and incorporate the previous belief state. This is interesting as it shows that the information previously given in multi-turn is still usable, but needs to be called explicitly.
  \item \textbf{Basic Reasoning Failure:} ChatGPT's response tends to be wrong if the query introduces a basic level of reasoning such as when it is asked for ``recommendation for restaurants with European food'' (ChatGPT has to filter the types of cuisine which are based on countries) or ``recommendation for restaurants with a rating of 3 or higher'' (ChatGPT needs to understand rating 3, 4 and 5). Even with a basic knowledge base, ChatGPT fails to answer correctly 66\% of the time.
  \item \textbf{Extrinsic Hallucination:} ChatGPT tends to generate hallucinated information beyond the given knowledge. This is especially harmful in TOD as ChatGPT will sometimes hallucinate some prices for hotel booking, or availability for restaurants. 
\end{itemize}

We provide the example for the modular and unified approaches for Task-Oriented Dialogue in \Cref{tab:tod-modular} and \Cref{tab:tod-unified}, respectively.

\clearpage

\begin{table*}[!ht]
\small
\centering
\resizebox{\linewidth}{!}{%
\begin{tabular}{p{0.13\linewidth} | p{0.1\linewidth} | p{0.77\linewidth}}
\toprule
\textbf{Task} & \textbf{Key} & \textbf{Text Content} \\ \midrule
\multirow{7}{*}{\textbf{\begin{tabular}[c]{@{}l@{}}Dialogue State\\Tracking\end{tabular}}} & \multicolumn{1}{l|}{Prompt} &
  \begin{tabular}[c]{p{\linewidth}}Give the dialogue state of the last utterance in the following dialogue in the form of 'STATE: {Domain-Intent: [Slot, Possible value], ...} (for example: STATE: {Hotel-Inform: ['area', 'centre']}) by using the following pre-defined slots and possible values:\\
\\
Intents: Request, Inform, general-thank, general-bye\\
Domain: hotel, Slots: pricerange, Possible values: ['expensive', 'cheap', 'moderate']\\
Domain: hotel, Slots: type, Possible values: ['guesthouse', 'hotel']\\
Domain: hotel, Slots: parking, Possible values: ['free', 'no', 'yes']\\
Domain: hotel, Slots: bookday, Possible values: ['monday', 'tuesday', 'wednesday', 'thursday', 'friday', 'saturday', 'sunday']\\
Domain: hotel, Slots: bookpeople, Possible values: ['1', '2', '3', '4', '5', '6', '7', '8']\\
Domain: hotel, Slots: bookstay, Possible values: ['1', '2', '3', '4', '5', '6', '7', '8']\\
Domain: hotel, Slots: stars, Possible values: ['0', '1', '2', '3', '4', '5']\\
Domain: hotel, Slots: internet, Possible values: ['free', 'no', 'yes']\\
Domain: hotel, Slots: name, Possible values: []\\
Domain: hotel, Slots: area, Possible values: ['centre', 'east', 'north', 'south', 'west']\\
Domain: hotel, Slots: address, Possible values: [?]\\
Domain: hotel, Slots: phone, Possible values: [?]\\
Domain: hotel, Slots: postcode, Possible values: [?]\\
Domain: hotel, Slots: ref, Possible values: [?]\\
\\
USER: I need a five starts hotel close to a mall and main restaurants. The hotel should include free wifi in the room.\\
STATE : {'Hotel-Inform': [['internet', 'yes'], ['stars', '5']]}\\
SYSTEM: I'm sorry, but there are no 5-star hotels anywhere in Cambridge. Would you like to broaden your criteria?\\
USER: How about something with 4 stars and something expensive.\\
STATE : {'Hotel-Inform': [['pricerange', 'expensive'], ['stars', '4']]}\\
SYSTEM: I have 2 options for you - The Huntingdon Marriott on the west side, and the University Arms in the center. Are you interested in one of these?\\
USER: Maybe. Is either one a 4 star hotel? If so, I'd like to book a room for 4 nights.\\
STATE : {'Hotel-Inform': [['bookstay', '4'], ['stars', '4']]}\\
SYSTEM: Both are 4 stars. How many days would you be staying, and how many people are in your party? I can see if something is available.\\
USER: I need a reservation for 4 nights starting Saturday.\end{tabular} \\ \cmidrule{2-3}
 & \multicolumn{1}{l|}{Gold} &
  \begin{tabular}[c]{p{\linewidth}}STATE : {'Hotel-Inform': [['bookday', 'saturday'], ['stars', '4']]}\end{tabular} \\ \cmidrule{2-3}
 & \multicolumn{1}{l|}{\begin{tabular}[c]{@{}l@{}}ChatGPT\\ Generation\end{tabular}} &
  \begin{tabular}[c]{p{\linewidth}}STATE: {'Hotel-Inform': [['bookday', 'saturday'], ['bookstay', '4'], ['stars', '4']]}\end{tabular} \\ \bottomrule
  
  \multirow{7}{*}{\textbf{\begin{tabular}[c]{@{}l@{}}Response \\Generation \\using Actions\end{tabular}}} & \multicolumn{1}{l|}{Prompt} &
  \begin{tabular}[c]{p{\linewidth}}Continue the dialogue as a task-oriented dialogue system called SYSTEM. The answer of SYSTEM should follow the ACTION provided next while answering the USER's last utterance:\\
\\
ACTION: {'Restaurant-Inform': [['phone', '01223316074'], ['pricerange', 'cheap']]}\\
\\
USER: Where is a 4 star hotel located in North Cambridge?\\
SYSTEM: I have several options, almost all of which are moderate guesthouses. I do have one cheap one, too. Would you like to book a room?\\
USER: Sure , that could be nice\\
SYSTEM: OK, how many are in your party, what day will you arrive, and how many nights will you be staying?\\
USER: I actually don't need reservations I just need the phone number, price range.
\end{tabular} \\ \cmidrule{2-3}
 & \multicolumn{1}{l|}{Gold} &
  \begin{tabular}[c]{p{\linewidth}}SYSTEM: As I mentioned it is cheap and the phone number is 01223316074.\end{tabular} \\ \cmidrule{2-3}
 & \multicolumn{1}{l|}{\begin{tabular}[c]{@{}l@{}}ChatGPT\\ Generation\end{tabular}} &
  \begin{tabular}[c]{p{\linewidth}}SYSTEM: The phone number for the 4-star hotel located in North Cambridge is 01223316074 and it is in the cheap price range.\end{tabular} \\ \bottomrule
\end{tabular}}
\caption{Examples of modular Task-Oriented Dialogue using ChatGPT: dialogue state tracking and response generation}
\label{tab:tod-modular}
\end{table*}

\begin{table*}[!ht]
\small
\centering
\resizebox{\linewidth}{!}{%
\begin{tabular}{p{0.13\linewidth} | p{0.1\linewidth} | p{0.77\linewidth}}
\toprule
\textbf{Task} & \textbf{Key} & \textbf{Text Content} \\ \midrule
  \multirow{25}{*}{\textbf{\begin{tabular}[c]{@{}l@{}}Multi-turn \\unified approach\end{tabular}}} & \multicolumn{1}{l|}{Prompt} &
  \begin{tabular}[c]{p{\linewidth}}Use the following knowledge base to complete the task of ``recommending a restaurant'' by continuing the conversation as a task-oriented dialogue system:\\
 \\
 Restaurant: Mama Julia, Food: French, Price: Expensive, Location: 7th street, Rating: 5 \\
 Restaurant: Papa John, Food: Italian, Price: Moderate, Location: 3rd street, Rating: 4 \\
 Restaurant: The Crossroad, Food: Morocco, Price: Moderate, Location: Downtown, Rating: 2 \\
 Restaurant: Tacos City, Food: Mexian, Price: Cheap, Location: Center, Rating: 1 \\
 Restaurant: Golden Rice Bowl, Food: Chinese, Price: Cheap, Location: 3rd district, Rating: 3 \\
 Restaurant: Veggie Garden, Food: Chinese, Price: Expensive, Location: Town Hall, Rating: 4 \\
 Restaurant: Pizza House, Food: Italian, Price: Moderate, Location: 3rd street, Rating: 2 \\
 Restaurant: The Palace, Food: Vietnamese, Price: Expensive, Location: Hotel Grandview, Rating: 5 \\
 \\
 Hi, I'm looking for a restaurant\end{tabular} \\ \cmidrule{2-3}
& \multicolumn{1}{l|}{ChatGPT} &
  \begin{tabular}[c]{p{\linewidth}} Sure, what type of food are you in the mood for? \end{tabular} \\ \cmidrule{2-3}
 & \multicolumn{1}{l|}{\begin{tabular}[c]{@{}l@{}}Prompt\end{tabular}} &
  \begin{tabular}[c]{p{\linewidth}} I only want restaurants rated 3 or higher. Any suggestions? \end{tabular} \\ \cmidrule{2-3}
& \multicolumn{1}{l|}{ChatGPT} &
  \begin{tabular}[c]{p{\linewidth}} Here are some restaurants that have a rating of 3 or higher: Mama Julia (French, Expensive, 5 rating), Papa John (Italian, Moderate, 4 rating), Golden Rice Bowl (Chinese, Cheap, 3 rating), Veggie Garden (Chinese, Expensive, 4 rating), and The Palace (Vietnamese, Expensive, 5 rating). Which one would you like to know more about? \end{tabular} \\ \cmidrule{2-3}
 & \multicolumn{1}{l|}{\begin{tabular}[c]{@{}l@{}}Prompt\end{tabular}} &
  \begin{tabular}[c]{p{\linewidth}}What food does The Palace serve? \end{tabular} \\ \cmidrule{2-3}
& \multicolumn{1}{l|}{ChatGPT} &
  \begin{tabular}[c]{p{\linewidth}} The Palace serves Vietnamese cuisine. \end{tabular} \\ \cmidrule{2-3}
 & \multicolumn{1}{l|}{\begin{tabular}[c]{@{}l@{}}Prompt\end{tabular}} &
  \begin{tabular}[c]{p{\linewidth}} Is Taxos City expensive to eat at? \end{tabular} \\ \cmidrule{2-3}
& \multicolumn{1}{l|}{ChatGPT} &
  \begin{tabular}[c]{p{\linewidth}} Tacos City is considered as cheap restaurant. \end{tabular} \\ \cmidrule{2-3}
 & \multicolumn{1}{l|}{\begin{tabular}[c]{@{}l@{}}Prompt\end{tabular}} &
  \begin{tabular}[c]{p{\linewidth}} I want to eat Italian food, any recommendations? \end{tabular} \\ \cmidrule{2-3}
& \multicolumn{1}{l|}{ChatGPT} &
  \begin{tabular}[c]{p{\linewidth}} Based on the information provided in the knowledge base, I would recommend Papa John and Pizza House for Italian food. Papa John is rated 4 out of 5 and is classified as ``Moderate'' in terms of price, while Pizza House is rated 2 out of 5 and is also classified as ``Moderate''. Papa John is located on 3rd street, while Pizza House is located on 3rd street as well. \end{tabular} \\ \cmidrule{2-3}
 & \multicolumn{1}{l|}{\begin{tabular}[c]{@{}l@{}}Prompt\end{tabular}} &
  \begin{tabular}[c]{p{\linewidth}} Did you forget I only wanted restaurants rated 3 or higher? \end{tabular} \\ \cmidrule{2-3}
& \multicolumn{1}{l|}{ChatGPT} &
  \begin{tabular}[c]{p{\linewidth}} I apologize, I must have misunderstood your previous request. Based on the knowledge base provided, The only Italian restaurant that is rated 3 or higher is Papa John which is rated 4 out of 5 and is located on 3rd street. It is classified as ``Moderate'' in terms of price. \end{tabular} \\ \cmidrule{2-3}
& \multicolumn{1}{l|}{Prompt} &
  \begin{tabular}[c]{p{\linewidth}} Great thanks for your help. \end{tabular} \\ \cmidrule{2-3}
& \multicolumn{1}{l|}{ChatGPT} &
  \begin{tabular}[c]{p{\linewidth}} You're welcome! I hope you enjoy your meal at Papa John. If you have any other questions or need further assistance, feel free to ask. \end{tabular} \\ \bottomrule
\end{tabular}}
\caption{Example of multi-turn unified approach for Task-Oriented Dialogue using ChatGPT}
\label{tab:tod-unified}
\end{table*}

% ==================================================================
\newpage\newpage
\begin{table*}[!ht]
    \begin{adjustbox}{width=\linewidth,totalheight={\textheight},keepaspectratio}
    \centering
    \begin{tabular}{p{0.4\linewidth}|p{0.3\linewidth}|p{0.3\linewidth}}
        \toprule
        \multicolumn{1}{c|}{\textbf{ChatGPT}} & \multicolumn{1}{c|}{\textbf{InstructGPT}} & \multicolumn{1}{c}{\textbf{text-davinci-003}} \\
        \midrule
        % The language of the text appears to be a variation of the Bugis language spoken in Indonesia. & It appears to be a made-up language. & The language of the text is Indonesian. \\
        The language of the text appears to be a variant of the \valid{Bugis language spoken in Indonesia}. & The language of the text is the \error{Sasak language}, spoken in \error{Lombok}, Indonesia. & The text is written in \valid{Buginese}. \\ \midrule
        I am sorry, I do not recognize the language of the text. & The language of the text is \error{Koyukon Athabascan}. & The text is in the \error{Balinese language}. \\ \midrule
        The language of the text appears to be a \error{dialect} of the Indonesian language. & The language of the text is \error{Indonesian}. & The language of the text is \error{Indonesian}. \\ \bottomrule
    \end{tabular}
    \end{adjustbox}
    \caption{Example of \textbf{Buginese} language identification response from ChatGPT, InstructGPT, and text-davinci-003.}
    \label{tab:chat-gpt-nlu}
\end{table*}

\clearpage
\section{ChatGPT on Multilinguality}
\label{app:multilinguality}

We present the statistics of language under study in~\Cref{tab:cc_distribution}. In the following section, we provide the insights that we find during our experiment in exploring multilingual capability of ChatGPT.

\begin{table}[]
\centering
    \begin{adjustbox}{width=\linewidth,totalheight={\textheight},keepaspectratio}
    \begin{tabular}{c|c|c|c}
    \toprule
    \textbf{Language} & \textbf{\#Speakers} & \textbf{CC Size (\%)} & \multicolumn{1}{c}{\textbf{\begin{tabular}[c]{@{}c@{}}Language\\  Category\end{tabular}}} \\ \midrule
    English (eng) & 1.452B  & 46.320 & HRL \\
    Chinese (zho) & 1.118B & 4.837 & HRL   \\
    French (fra) & 235M & 4.604 & HRL   \\
    Indonesian (ind) & 199M & 0.781 & MRL \\
    Korean (kor) & 81.7M & 0.679 & MRL \\
    Javanese (jav) & 68.3M & 0.002 & LRL \\
    Sundanese (sun) & 32.4M & 0.001 & LRL \\ 
    Buginese (bug) & 5M & 0.000 & X-LRL \\ \bottomrule
    \end{tabular}
    \end{adjustbox}
    \caption{The statistics of languages used in our language disparity experiment. 
    \textbf{HRL}, \textbf{MRL}, \textbf{LRL}, \textbf{X-LRL} denote high-, medium-, low-, extremely low-resourced language respectively. 
    % \textbf{HRL} denotes high-resourced language, \textbf{MRL} denotes medium-resourced language, \textbf{LRL} denotes low-resourced language, \textbf{X-LRL} denotes extremely low-resourced language. 
    % \#Speakers is collected from the total speakers (L1+L2) information on Wikipedia.\footnotemark
    }
    \label{tab:cc_distribution}
\end{table}

\paragraph{ChatGPT understands LRL sentences but fails to identify their language} As shown in Table~\ref{tab:nlu_result_app}, ChatGPT correctly classifies the languages for English and Indonesian 100\% of the time. While for the language identification for Javenese and Buginese, ChatGPT either misclassifies the samples as other languages or is unable to determine the language. 
% ChatGPT misclassifies the samples mostly as Indonesian, despite having various dissimilarities across languages~\cite {grimes2000ethnologue,lewis2009ethnologue,cohn2014local,eberhard2021ethnologue,aji-etal-2022-one,cahyawijaya2022nusacrowd}. 
Nevertheless, ChatGPT performance on the sentiment analysis in Javanese is only slightly lower compared to English and Indonesian which suggests that ChatGPT can understand the semantic meaning of sentences in low-resource languages without having the knowledge to identify the language itself. This limitation of language identification in LMs aligns with the result from BIG-bench~\cite{bigbench}.

\paragraph{ChatGPT displays better human-preferred responses} As shown in \Cref{tab:chat-gpt-nlu}, ChatGPT lets the user know that its prediction is uncertain when it does not completely understand the language and also provides broader information regarding the language, such as location and tribe of which the predicted language is spoken. This fact provides evidence regarding the benefit of using the RLHF approach compared to other training approaches for aligning LLMs with human preferences. 
% \pascale{this is an important point that should have been explored beyond multilinguality.}

\begin{table}[]
    \centering
    \resizebox{0.85\linewidth}{!}{
    \begin{tabular}{c|c|c}
        \toprule
        \textbf{Language} & \textbf{SA Acc.} & \textbf{LID Acc.}  \\ \midrule
        English & 84\% & 100\% \\
        Indonesian & 80\% & 100\% \\
        Javanese & 78\% & 0\% \\
        Buginese & 56\% & 12\% \\
        \bottomrule 
    \end{tabular}
    }
    \caption{Accuracy of ChatGPT on Sentiment Analysis (SA) and Language Identification (LID) tasks.}
    \label{tab:nlu_result_app}
\end{table}

\paragraph{ChatGPT understands non-Latin scripts better than it can generate them}

Despite being high-resource and medium-resource languages, the translation from English to Chinese and Korean is much inferior to the other languages with Latin scripts, i.e., French or Indonesian. Similarly, prior works focusing on transliteration~\cite{chau-smith-2021-specializing,muller-etal-2021-unseen} have shown the effectiveness of utilizing Latin scripts over other scripts, e.g., Cyrillic, Georgian, Arabic, etc, especially for low-resource languages. Interestingly, this problem of using non-Latin scripts is less severe for translation from Chinese and Korean to English, which suggests that ChatGPT can better neutralize the effect of non-Latin scripts as source languages~\cite{wan2022fairness}, but it still lacks the ability to generate non-Latin script languages.

% ===========MULTIMODAL======================================
\section{Multimodality: Flag Drawing Task}
\label{app:mm-flag-appendix}

\paragraph{Task Formulation} We uniformly collect 50 national flags from different continents and conduct the flag-drawing task on ChatGPT. 
The flag-drawing task contains three steps:
\begin{enumerate}
    \item Ask ChatGPT to illustrate the appearance of the flag using the prompt ``\texttt{Describe how the <NATION> flag looks like}''.
    \item Based on the description, ask ChatGPT to generate the SVG code of that flag by prompting ``\texttt{Generate a code snippet to represent that flag in SVG format}''.
    \item If the generated image contains errors, we iteratively ask ChatGPT to fix them. 
\end{enumerate}
% Firstly, we ask ChatGPT to illustrate the appearance of the flag using the prompt ``\texttt{Describe how the <NATION> flag looks like}''. Next, based on the description, we ask ChatGPT to generate the SVG code of that flag by prompting ``\texttt{Generate a code snippet to represent that flag in SVG format}''. Finally, if the generated image contains errors, we iteratively ask ChatGPT to fix them. 
There are four types of evaluation criteria: 1) layout 2) color 3) missing components 4) shape/size. In each round of fixing, we ask ChatGPT to revise only one type of error with the prompt ``\texttt{<ERROR DESCRIPTION>. Revise the image}''. We terminate the conversation once the generated flag becomes perfect or we have already passed two rounds of fixing.

\begin{figure*}[!ht]
    \centering
    % \fbox{
    \includegraphics[width=\linewidth, trim={0 0 0 0}, clip]{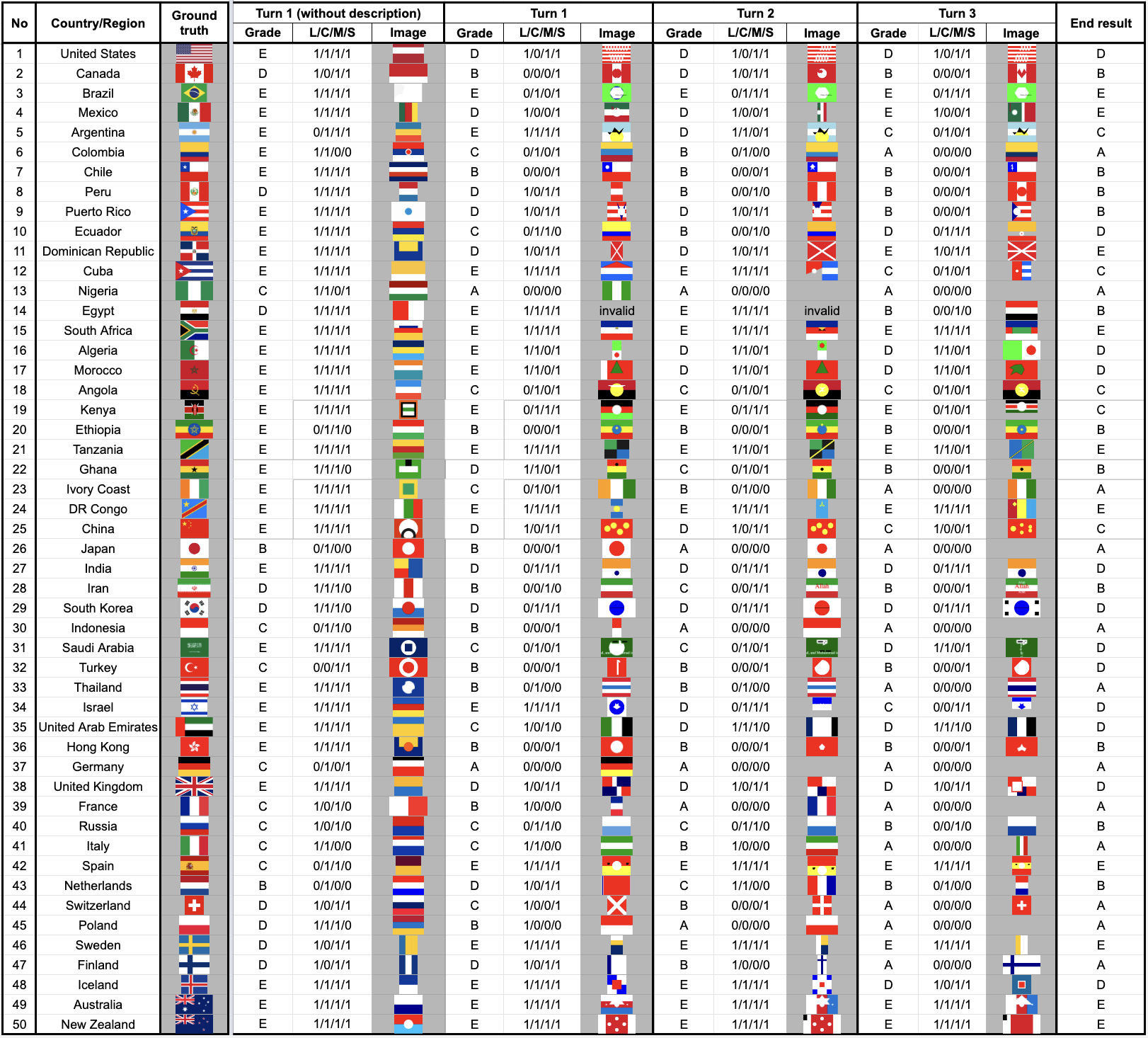}
    % }
    \caption{Complete results of the flag drawing task. Multi-turn refinement allows ChatGPT to generate a more similar image to the ground truth image.}
    \label{fig:flag-appendix}
\end{figure*}

\paragraph{Evaluation} 
The generated flag images are evaluated by the aforementioned four error types as criteria. We further assess the image quality with five grades, A $\sim$ E, which indicate zero to four (or above) errors. We assign grades to each round so that we can assess the number of improvements and degradation through conversational interactions (post-editing). The full results are shown in Figure~\ref{fig:flag-appendix}.

% ========== REASONING APPENDIX

\section{Details for Reasoning Evaluations}
\label{sec:reasoning-appendix}
\Cref{tab:reasoning_cat} shows the categories of reasoning that are evaluated in this paper as well as corresponding datasets. The following section introduces each of the categories and detailed experimental setup and/or analysis.

\begin{table}[!ht]
\centering
\resizebox{\columnwidth}{!}{%
\begin{tabular}{cl}
\toprule
\textbf{Categories} & \multicolumn{1}{c}{\textbf{Dataset}} \\ \midrule
Deductive & \begin{tabular}[c]{@{}l@{}} EntailmentBank~\citep{dalvi2021explaining}\\ bAbI (task 15)~\citep{weston2016towards}\end{tabular} \\\midrule
Inductive & \begin{tabular}[c]{@{}l@{}}CLUTRR~\citep{sinha2019clutrr}\\ bAbI (task16)~\citep{weston2016towards}\end{tabular} \\\midrule
Abductive & $\alpha$NLI ~\citep{bhagavatula2020abductive} \\\midrule
Temporal & Timedial~\citep{qin-etal-2021-timedial} \\\midrule
Spatial & \begin{tabular}[c]{@{}l@{}}SpartQA~\citep{mirzaee-etal-2021-spartqa}\\ StepGame~\citep{Shi_Zhang_Lipani_2022}\end{tabular} \\\midrule
Mathematical & Math~\citep{saxton2018analysing}\\ \midrule
Commonsense & \begin{tabular}[c]{@{}l@{}} CommonsenseQA~\citep{talmor2018commonsenseqa}\\PiQA~\citep{bisk2020piqa}\\Pep-3k~\citep{wang2018modeling}\end{tabular} \\\midrule
Causal & E-Care~\citep{du-etal-2022-e}\\\midrule
Multi-hop & HotpotQA~\citep{yang2018hotpotqa} \\ \midrule
Analogical & Letter string
analogies~\citep{webb2022emergent} \\ 
\bottomrule
\end{tabular}}
\caption{Reasoning categories and corresponding datasets used to evaluate ChatGPT in this work.}
\label{tab:reasoning_cat}
\end{table}

\subsection{Logical Reasoning}
\label{sec:logical-reasoning-appendix}

Inductive and deductive are categorized by ``a degree to which the premise supports the conclusion'' based on logic and philosophy \citep{qiao2022reasoning, rogers2022qa, sep-logic-inductive}. Inductive reasoning is based on ``observations or evidence'' while deductive is based on ``truth of the premises'' (i.e., necessarily true inference) \citep{douven_2017}. Another way to categorize is based on the ``direction of reasoning'' -- deductive is from premise to conclusion while abductive is from conclusion to the most probable premise that supports the conclusion \citep{walton2014abductive}.

Inductive and deductive reasoning are common forms of logical reasoning that are categorized by ``a degree to which the premise supports the conclusion'' based on logic and philosophy \cite{qiao2022reasoning, rogers2022qa, sep-logic-inductive}. Deductive reasoning involves processes of driving specific conclusions based on more general premises. On the contrary inductive reasoning involves specific observation of patterns, processing them on increasingly abstract cycles of hypothetico-deductive reasoning to draw a more general conclusion \citep{lawson2005role}. Comparing the two types of reasoning, deduction requires less ``guessing'' from the perspective of ChatGPT, as induction requires figuring out rules~\citep{rogers2022qa}. The former can be viewed as top-down while the latter is bottom-up.
%  # [HERE WAS SPATIAL TABLE]

\subsubsection{Deductive vs. Inductive Reasoning}
Deductive reasoning involves processes of driving specific conclusions based on \textit{more general premises}. On the contrary, inductive reasoning involves \textit{specific observation of patterns}, processing them on increasingly abstract cycles of hypothetico-deductive reasoning to draw a more general conclusion \citep{lawson2005role}. Comparing the two types of reasoning, deduction requires less ``guessing'' from the perspective of ChatGPT, as induction requires figuring out rules~\citep{rogers2022qa}. The former can be viewed as top-down while the latter is bottom-up.

We explore ChatGPT's ability of inductive and deductive reasoning in two different levels: 1) basic and 2) advanced. Basic-level tasks are the prerequisites to probe reasoning. While solving these tasks does not necessarily indicate full reasoning capability, if ChatGPT fails on any of these tasks, then there are likely real-world tasks that it will fail on too if they require similar reasoning mechanisms. Consequently, the advanced-level tasks are there to probe those capabilities in real-world tasks where the noises are present, and solving them requires a more systematic generalization. Additionally, we choose tasks that do not require or are dependent on external knowledge and the solution could be only derived by premises to focus on dissecting the capability of each reasoning mechanism.

\begin{table}[]
\centering
\resizebox{\columnwidth}{!}{%
\begin{tabular}{ccc}
\toprule
\multicolumn{3}{c}{\textbf{Deductive Reasoning Tasks}} \\ \midrule
\textbf{bAbI - task 15} & \textbf{\begin{tabular}[c]{@{}c@{}}bAbI - task 15\\ (prompt engineered)\end{tabular}} & \textbf{EntailmentBank} \\ \midrule
19/30            & 28/30            & 28/30            \\ \midrule \midrule
\multicolumn{3}{c}{\textbf{Inductive Reasoning Tasks}} \\ \midrule
\textbf{bAbI - task16}  & \textbf{\begin{tabular}[c]{@{}c@{}}bAbI - task 16\\ (prompt engineered)\end{tabular}} & \textbf{CLUTRR}         \\ \midrule
0/30             & 20/30            & 13/30 \\ \bottomrule  
\end{tabular}%
}
\caption{Inductive vs. Deductive Reasoning. ChatGPT performs better deduction rather than induction. Engineering the prompt to explicitly ask ChatGPT to do reasonable inference improves its reasoning capability. The scores are in accuracy over tested samples.}
\label{tab:deductiveinductive}
\end{table}
% \pascale{need to describe also abductive reasoning}
% --- og reasoning end
\begin{table*}[!]
\small
\centering
\resizebox{\linewidth}{!}{%
\begin{tabular}{p{0.14\linewidth}|
                p{0.635\linewidth}|
                p{0.15\linewidth}|
                p{0.07\linewidth}|
                p{0.005\linewidth}}
% \begin{tabular}{llllc}
\toprule
\multicolumn{1}{c|}{\textbf{Task}} &
  \multicolumn{1}{c|}{\textbf{Prompt}} &
  \multicolumn{1}{c|}{\textbf{ChatGPT answer}} &
  \multicolumn{1}{c|}{\textbf{Gold}} &
  \multicolumn{1}{c}{\textbf{T/F}} \\ \midrule \midrule
\multicolumn{5}{c}{\textbf{Deductive Reasoning Tasks}} \\ \midrule
\textbf{bAbI (task 15)} &
  \begin{tabular}[c]{@{}l@{}}1 Wolves are afraid of mice.\\2 Sheep are afraid of mice.\\3 Winona is a sheep.\\4 Mice are afraid of cats.\\5 Cats are afraid of wolves.\\6 Jessica is a mouse.\\7 Emily is a cat.\\8 Gertrude is a wolf.\\10 What is winona afraid of?\end{tabular} &
  Winona, being a sheep, is afraid of mice. &
  mouse &
  T \\ \midrule
\textbf{\begin{tabular}[c]{@{}l@{}}bAbI (task 15)\\ (prompt \\engineered)\end{tabular}} &
  \mline{\begin{tabular}[c]{@{}l@{}}Given facts: Wolves are afraid of mice. Sheep are afraid of mice. Winona is a \\sheep. Mice are afraid of cats. Cats are afraid of wolves. Jessica is a mouse. \\Emily is a cat. Gertrude is a wolf. \\ \\Based on the given facts above, do a reasonable inference on this question using \\deductive reasoning: \\What is winona afraid of?\end{tabular}} &
  \mline{Based on the given facts, it can be inferred that Winona is afraid of mice.}&
  \mline{mouse} &
  \mline{T} \\ \midrule
\mline{\shortstack[l]{\textbf{Entailment-}\\\textbf{Bank}}} &
  \mline{earth is a kind of planet. a planet is a kind of celestial object / celestial body. earth is located in the milky way galaxy. Which object is a planet found in the Milky Way Galaxy?} &
  \mline{Earth is a planet located in the Milky Way galaxy.} &
  \mline{Earth} &
  \mline{T} \\ \midrule \midrule
\multicolumn{5}{c}{\textbf{Inductive Reasoning Tasks}} \\ \midrule
\textbf{bAbI (task 16)} &
  \mline{\begin{tabular}[c]{@{}l@{}}1 Bernhard is a swan.\\ 2 Greg is a frog.\\ 3 Brian is a swan.\\ 4 Julius is a lion.\\ 5 Greg is gray.\\ 6 Julius is yellow.\\ 7 Lily is a lion.\\ 8 Lily is green.\\ 9 Brian is yellow.\\ 10 What color is Bernhard?\end{tabular}} &
  \mline{It is not specified in the given information what color Bernhard is.} &
  \mline{yellow} &
  \mline{F} \\ \midrule
\textbf{\begin{tabular}[c]{@{}l@{}}bAbI (task 16)\\ (prompt \\engineered)\end{tabular}} &
  \mline{\begin{tabular}[c]{@{}l@{}}Given facts: Bernhard is a swan. Greg is a frog. Brian is a swan. Julius is a lion. \\Greg is gray. Julius is yellow. Lily is a lion. Lily is green. Brian is yellow.\\ \\ The most recent fact is the correct fact.\\ \\ Based on the given facts above, do a reasonable inference on this question using \\inductive reasoning: \\ What color is Bernhard?\end{tabular}} &
  \mline{Based on the given facts, it can be inferred that Bernhard is yellow, as Brian is also stated to be yellow.} &
  \mline{yellow} &
  \mline{T} \\ \midrule
\multirow{4}{*}{\textbf{CLUTRR}} &
  \mline{{[Jason]} and his wife {[Gabrielle]} went to the beach to watch the fireworks on the 4th of July. {[Jason]} and his daughter {[Alma]} took a day off school to go to the zoo... Who is Alma to Gabrielle?} &
  \mline{Alma is the daughter of Jason and Gabrielle.} &
  \mline{daughter} &
  \mline{T} \\ \cmidrule(lr){2-5}
 &
  \mline{{[Jason]} took his grandson {[Donald]} fishing. {[Russell]} enjoys going fishing with his brother. His name is {[Donald]}... Who is Russell to Jason?} &
  \mline{Russell is the brother of Jason.} &
  \mline{grandson} &
  \mline{F} \\ \midrule \bottomrule
\end{tabular}%
}
\caption{Prompting samples on deductive and inductive reasoning tasks. ChatGPT is performing better deduction rather than induction. On both types of reasoning, when ChatGPT is explicitly asked to do reasonable inferences, its ability for reasoning increases. Additionally, it also makes frequent mistakes regarding the grandson's kinship.}
\label{tab:deductiveinductivesample}
\end{table*}

\begin{table*}[!]
\small
\centering
\resizebox{\linewidth}{!}{%
\begin{tabular}{p{0.12\linewidth}|
                p{0.05\linewidth}|
                p{0.4\linewidth}|
                p{0.26\linewidth}|
                p{0.07\linewidth}|
                p{0.005\linewidth}}
\toprule
\multicolumn{6}{c}{\textbf{StepGame (Basic) Breakdown Analaysis}} \\ \midrule
\multicolumn{1}{c}{} &
  Result &\multicolumn{1}{c}{Example} &
  \multicolumn{1}{c}{ChatGPT answer} &Gold &T/F \\ \midrule
Clock-position &
  5/20 &
  G is at Y's 6 o'clock. What is the relation of the agent Y to the agent G?&
  The agent Y is to the right of the agent G. &
  Above &
  F \\\midrule
\mline{Basic Cardinal} &
  17/20 &
  D and K are parallel, and D is under K. What is the relation of the agent K to the agent D? &
  The spatial relation of the agent K to the agent D is that K is above D. &
  Above &
  \mline{T} \\\midrule
\mline{Diagonal} &
  11/20 &
  W presents lower left to I. What is the relation of the agent I to the agent W? &
  The relation of the agent I to the agent W is lower-left. &
  Upper-Right &
  \mline{F} \\ \bottomrule
\end{tabular}}
\caption{A provided illustration to help the readers to understand each comparison between the categories (not the actual prompts). We provide the options to ChatGPT as: \texttt{Choose from: left, right, above, below, lower-left, lower-right, upper-left, upper-right}.}
\label{tab:spatialbreakdown}
\end{table*}

% ==================================================================
% \textbf{ChatGPT is a lazy reasoner that suffers more with induction} Interestingly, when ChatGPT was asked to answer a question given premises without any prompt engineering, it performs poorly in inductive reasoning (0 out of 30) while it achieves much better performance in deductive (19 of 30). However, when ChatGPT is explicitly asked for reasonable inference inductive reasoning increases to 20 out of 30. Yet, it is still not as good as in deduction.

\paragraph{Results \& Detailed Analysis}
ChatGPT answers \textit{``It is not specified what <attribute> <entity> is.''} for most of the time when it was asked a question requiring inductive reasoning. However, when ChatGPT is explicitly asked for reasonable inference with a prompt \textit{``Based on the given facts, do a reasonable inference on this question using inductive reasoning:''}, its ability for inductive reasoning increases. Yet, it is still not as good as in deduction as the same prompt engineering also helps increases its ability for deductive reasoning.

We could derive similar insight as ChatGPT only correctly answered for half of the time while it could make inferences deductively well for 90\% of the time. CLUTRR~\cite{sinha2019clutrr} requires induction on extracting relations between entities, and in the ChatGPT responses, it often asks for more information to make inferences. An interesting finding along with CLUTRR was that ChatGPT can't differentiate son and grandson but can differentiate daughter and granddaughter when it induces the logical rules governing kinship relationships. We show all performances in Table \ref{tab:deductiveinductive} and some of the prompting samples in Table \ref{tab:deductiveinductivesample}. We follow \cite{qiao2022reasoning} categorization on the deductive and inductive reasoning datasets, but we only use the QA part of EntailmentBank, that the authors took from ARC dataset \cite{clark2018arc}, as we aim to test for reasoning capability. Regarding EntailmentBank, it might trigger the universe-related knowledge out of ChatGPT, which could help the model to derive the correct answer, although the test set is designed to test deductive reasoning skills. One of the future explorations would be with checking the rationale of ChatGPT as a follow-up question.

\subsubsection{Abductive Reasoning}

% \begin{table}[hbt!]
% \centering
% \begin{adjustbox}{width=0.1\linewidth,totalheight={\textheight},keepaspectratio}
%     \begin{tabular}{c}
%     \toprule
%     % \multicolumn{3}{c}{\textbf{Abductive Reasoning Tasks}} \\
%     $\alpha$NLI \\ \midrule
%     26/30 \\ \bottomrule
%     \end{tabular}
%     \end{adjustbox}
%     \caption{Abductive reasoning. ChatGPT is performing good abductive reasoning capability.}
%     \label{tab:spatial}
% \end{table}

Abductive reasoning is the inference to the most plausible explanation given observations. For instance, ``if Jenny finds her house in a mess when she returns from work, and remembers that she left a window open, she can hypothesize that a thief broke into her house and caused the mess''~\footnote{An example provided by \citet{bhagavatula2020abductive}.}. We test ChatGPT's language-based abductive reasoning ability with 30 samples from $\alpha$NLI dataset \citep{bhagavatula2020abductive}, which requires the model to select the most plausible explanation given the conclusion. Based on our test, it could achieve 86.7\% (26 out of 30) accuracy.

\subsection{Non-textual Semantic Reasoning}
\label{sec:non-text-reasoning-appendix}

\paragraph{Mathematical reasoning}
Mathematical capabilities or numerical reasoning has been frequently mentioned to be lacking for LLMs, not only ChatGPT \citep{mathchatgpt}. \citeauthor{mathchatgpt} test ChatGPT's capability with publicly available datasets as well as the human-curated dataset, which consists of 728 prompts. The shared findings for ChatGPT's mathematical capabilities include 1) ChatGPT often understands the question but fails to provide correct solutions; 2) it shows inconsistent poor performance on graduate-level advanced mathematics; 3) it has a great ability to search for mathematical objects. \footnote{Refer to detailed findings in the original paper.} We also test separately on MATH dataset. Not surprisingly, it could only score 23.33\% (7/30) for the MATH dataset \citep{saxton2018analysing}, which tests mathematical reasoning. 

\paragraph{Temporal reasoning}
Temporal reasoning is mentioned a few times in the literature but is less common than others. It tests the understanding of the time duration of and the relation between events. For this category, we conduct experiments on the dataset TimeDial \citep{qin-etal-2021-timedial}, which solely requires temporal reasoning. We follow the format of the task in the BIG-bench benchmark ~\citep{bigbench}, which is multiple-choice (single correct answer),
Overall, ChatGPT correctly answers 86.67\% of the time (26/30), suggesting that it has a decent temporal reasoning ability. Also, compared to Chinchilla and Gopher which have the accuracy of 68.8\% and 50.9\% respectively, ChatGPT shows a promising improvement for LLMs in that aspect.

\begin{table}[t!]
\centering
\begin{adjustbox}{width=0.7\linewidth, keepaspectratio}
    \begin{tabular}{cc|cc}
    \toprule
    \multicolumn{4}{c}{\textbf{Spatial Reasoning Tasks}} \\
    Dataset& Total & Basic & Hard\\ \midrule
    StepGame& 26/60  & 19/30 & 7/30 \\ 
    SpartQA & 28/64 & 20/32 & 8/32  \\ \bottomrule
    \end{tabular}
    \end{adjustbox}
    \caption{Spatial reasoning ability of ChatGPT. Overall, ChatGPT falls short of the task.}
    \label{tab:spatial}
\end{table}

\paragraph{Spatial Reasoning}
Spatial reasoning is using an understanding of spatial relations among different objects and spaces. For spatial reasoning, we utilize two existing datasets: SpartQA \citep{mirzaee-etal-2021-spartqa} and StepGame \citep{Shi_Zhang_Lipani_2022}, which compose of story-question pairs about k relations of k+1 (where k is up to 10) entities written in natural language. ChatGPT is asked to answer spatial relations between two entities based on the provided descriptions of different entities. ChatGPT falls short of the spatial reasoning tasks, as shown in Table \ref{tab:spatial}, with overall success rates of 43.33\% for StepGame and 43.75\% for SpartQA. ChatGPT could only score 25\% on SpartQA (hard), which covers multiple spatial reasoning sub-types, and 23.33\% for stepGame (Hard) with k=9. ChatGPT could not provide any spatial relations but instead generated \textit{``It is not specified in the given description''}. Even with the fine-tuned models, as the number of relations (k) increases in context description, performance drops \citep{Shi_Zhang_Lipani_2022}. 

To understand spatial reasoning ability at a more elementary level, we test with less complicated examples from StepGame which we refer to as \textbf{StepGame (Basic)}. It does not involve multi-hop reasoning but purely spatial relation between two entities. (e.g, \textit{``C is sitting at the top position to Y. What is the relation of the agent Y to the agent C?''}). We test for basic spatial relations with 8 labels from StepGame \{left, right, above, below, lower-left, lower-right, upper-left, upper-right\}. When we test on StepGame (Basic), ChatGPT scores higher (63.33\%). 

We investigate the errors that it often fails to understand clock direction (e.g., ``W is at K's 3 o'clock'') and diagonal spatial relations. We further analyze the results by breaking down the test examples of StepGame (Basic) into two comparisons: i) types of directions (basic cardinal vs. diagonal) and ii) ways of spatial description for cardinal directions (basic cardinal\footnote{Those of which spatial relations are described with explicit vocabulary.} vs. clock-position cardinal). We take 20 more samples for each category (basic cardinal, diagonal, clock-position cardinal) and tested them as illustrated in Table \ref{tab:spatialbreakdown}.
% We further analyze by breaking test examples down into three categories 1) Cardianal 2) Dialognal 3) Clock-position description.

\begin{itemize}
  \item \textbf{ChatGPT poorly infers with clock-position description.}
  Although it is a simple cardinal direction, ChatGPT could only correctly answer for 5 samples (25\%), which is clearly poorer performance in comparison to performance with the basic cardinal description (17 correct answers). 
  \item \textbf{ChatGPT is worse at the diagonal position.} It correctly answers around half of the time (55\%), which is worse than basic cardinal points (85\%). Even with analysis from StepGame (Hard), among the correct 7 answers, there is only one diagonal direction that ChatGPT gets correctly while the others are all cardinal points. For those answers that require diagonal points, ChatGPT only could infer cardinal points for some examples.
\end{itemize}

\subsection{Commonsense Reasoning}
\label{sec:commonsense-reasoning-appendix}

\begin{table}[]
\centering
\begin{adjustbox}{width=0.85\linewidth,totalheight={\textheight},keepaspectratio}
    \begin{tabular}{ccc}
    \toprule
    \multicolumn{3}{c}{\textbf{Commonsense Reasoning Tasks}} \\
    CommonsenseQA & PiQA & Pep-3k (Hard) \\ \midrule
    27/30 & 25/30 & 28/30 \\ \bottomrule
    \end{tabular}
    \end{adjustbox}
    \caption{Commonsense reasoning ability of ChatGPT. ChatGPT shows good performance of commonsense reasoning capability on the three test data we test it on.}
    \label{tab:commonsense}
\end{table}

% % \pascale{need an introduction section on what is commonsense reasoning here!}

% Commonsense reasoning is the understanding and reasoning about everyday concepts and knowledge that most people are familiar with, to make judgments and predictions about new situations~\cite{storks2019commonsense}. Recent work has shown that LLMs perform impressively well on commonsense reasoning benchmarks~\cite{qiao2022reasoning, huang2022towards,bhargava2022commonsense}. However, ~\citeauthor{bhargava2022commonsense} also point out that the reasoning tasks underlying these benchmarks are still far from being solved, since most existing studies primarily report the performance of the models, without a detailed examination of the quality of the rationales produced. 

% \pascale{need more citations} \citeauthor{porada2021modeling} pointed out language models (LMs) often fail due to \textit{selectional preference}. For example, LMs infer that 'a person breathing' is plausible but `a dentist breathing' is not, because the word `dentist' rarely appears to be the nominal subject of the verb `breathe', although dentist is also a person -- this failure of conceptualizing is often seen as a lack of commonsense reasoning. 

To evaluate ChatGPT's capability on commonsense reasoning, we first test it on two widely used benchmark datasets CommonsenseQA~\cite{talmor2018commonsenseqa} and PiQA~\cite{bisk2020piqa}. CommonsenseQA focuses on general commonsense question answering such as \textit{``Where is a business restaurant likely to be located?''}, and PiQA is about physical commonsense reasoning: given a sentence such as \textit{``When boiling butter, when it’s ready, you can "}, the goal is to fill in the blank with one of two answer options, \textit{``Pour it onto a plate"} and \textit{``Pour it onto a jar"}. We use the validation split for both of the datasets since there are no labels provided on the test set that we retrieve. We also further probe ChatGPT by evaluating a more challenging commonsense reasoning dataset in a more comprehensive way. We use Pep-3k~\cite{wang2018modeling}, which requires the model to recognize plausible but possibly novel events, such as \textit{``man swallow paintball"}. Each instance in the Pep-3k is an s-v-o predicate, and the task is to judge if the predicate is plausible or not. But instead of evaluating ChatGPT's performance only based on the binary judgment, we also check if the answer contains relevant rationales (explanations) that lead to its judgment.

% This requires both commonsense knowledge and abstraction, which allows us to measure the commonsense reasoning ability of models. For quantitative results, ChatGPT achieves the performance of accuracy 28/30 and AUC 0.93, beating the State-of-the-Art result AUC 0.67 reported in ~\cite{porada2021modeling}. 

\begin{table}[]
\resizebox{\linewidth}{!}{%
\begin{tabular}{p{0.2\linewidth}|p{0.8\linewidth}}
\toprule
Prompt &
  Please judge if this predicate is (likely) plausible or implausible: ``knife chop rock'' \\ \midrule
ChatGPT Answer & It is \textbf{implausible} that a knife could chop through a rock. \valid{Knives are made to cut through softer materials} such as food, paper, or wood. \valid{Rocks are much harder than these materials} and a knife would not be able to make a dent in a rock, let alone chop it. \\ \bottomrule
\end{tabular}}
\caption{An example from Pep-3k~\cite{wang2018modeling} for commonsense reasoning of ChatGPT. We make the main answer \textbf{bold}, and highlight the explanation by green color.}
\label{tab:commonsene_ex}
\end{table}

For the Pep-3k samples, we prepend the s-v-o predicate with \textit{``Please judge if this predicate is (likely) plausible or implausible:"} to prompt ChatGPT. We show the results in Table~\ref{tab:commonsense}. As we see, ChatGPT performs quite well on the three datasets in terms of answer accuracy, which matches our anticipation. Furthermore, as we also check the rationales in ChatGPT's answer when evaluating Pep-3k samples, we can see that ChatGPT does quite well not only in terms of answer accuracy but also in generating reasonable reasoning procedures to support its answer. We show a concrete example in Table~\ref{tab:commonsene_ex}. As we can see, ChatGPT's answer explains well what kinds of materials are usually cut through with knives (i.e., food, paper, or wood). Then, it reasons why rocks cannot be chopped with a knife by explaining `rocks are much harder than these materials.'  While our findings are based on 30 samples from each dataset, we see the potential in ChatGPT's commonsense reasoning capability, and further large-scale investigation is worth exploring.

\subsection{Causal, Multi-Hop, and Analogical Reasoning}
\label{sec:other-reasoning}

\paragraph{Causal Reasoning}
Causal reasoning is the process of identifying the relationship between causes/actions and effects/changes (i.e., causality) \citep{thomason_2018,huang2022towards}. We test ChatGPT on 30 samples of human-annotated explainable CAusal REasoning dataset (E-CARE) \citep{du-etal-2022-e} and it could score 24 samples correctly (80\%). Note that our evaluation is mainly based on whether the model can make a judgment on correct causes or effects instead of its generated explanation of why the causation exists.

\paragraph{Multi-hop Reasoning}
To be able to reason over a larger context, a system has to perform multi-hop reasoning over more than one piece of information to arrive at the answer \cite{mavi2022survey}. We test ChatGPT's multi-hop reasoning capability on 30 samples of HotpotQA dataset \cite{yang2018hotpotqa} and we find that ChatGPT has difficulty performing with such capability, only answering 8 samples correctly, although the questions posed are only 2-hops. It is worth noting that ChatGPT oftentimes generates the answer in a short passage of explanations, thus we evaluate manually each of the ChatGPT responses to check its accuracy. This aligns with the findings that LLMs are also limited in several ways, and fail to produce accurate predictions due to their inability to accomplish complex reasoning, such as solving tasks that require multi-hop reasoning \cite{ott2023thoughtsource}.

\begin{table}[]
\centering
\resizebox{0.9\linewidth}{!}{%
\begin{tabular}{ccc}
\toprule
\textbf{Causal} & \textbf{Multi-hop} & \textbf{Analogical} \\ \midrule
E-CARE & HotpotQA & Letter string analogies \\ \midrule
24/30 & 8/30 & 30/30 \\ \bottomrule
\end{tabular}
}
\caption{Results for causal, multi-hop, and analogical reasoning. ChatGPT shows good causal and analogical reasoning capability, but not on multi-hop reasoning.}
\label{tab:other-reasoning}
\end{table}

\paragraph{Analogical Reasoning}
Analogical reasoning is a way of thinking that relies upon an analogy, comparing two or more objects or systems of objects \citep{bartha_2013} to drive a conclusion. We test with 30 samples from~\citet{webb2022emergent} and evaluate based on human evaluation, to see if the generated answer match with/contain the gold answer. ChatGPT could correctly answer all 30 examples, which may reveal that ChatGPT has a good capability in analogical reasoning skills. 

% ==================================================================

% ================ factualty and hallucination
\section{Details for Hallucination Evaluations}
\label{sec:hallu-appendix}

\begin{table*}[!]
\small
\centering
\resizebox{\linewidth}{!}{%
\begin{tabular}{p{0.13\linewidth} | p{0.1\linewidth} | p{0.77\linewidth}}
\toprule
\textbf{Task} & \textbf{Key} & \textbf{Text Content} \\ \midrule
\multirow{6}{*}{\textbf{\begin{tabular}[c]{@{}l@{}}Machine\\Translation\end{tabular}}} & \multicolumn{1}{l|}{Prompt} & 
\begin{tabular}[c]{p{\linewidth}}
(Javanese) Krajan Anyar Mesir kuno nggumun marang monumen-monumen leluhure, sing nalika iku tetep apik luwih seka sewu taun.
\end{tabular}\\ \cmidrule{2-3}
 & \multicolumn{1}{l|}{\begin{tabular}[c]{@{}l@{}}ChatGPT\\ Generation\end{tabular}} & \begin{tabular}[c]{p{\linewidth}}
The ancient Egyptian \error{Pharaohs} built beautiful monuments that are still well-preserved after thousands of years.
 \end{tabular}\\ \cmidrule{2-3}
 & \multicolumn{1}{l|}{Hallucination} &
  Extrinsic - New Kingdom ancient Egyptians marveled at their predecessors monuments, which were then well over a thousand year old. \\ \midrule

\multirow{7}{*}{\textbf{\begin{tabular}[c]{@{}l@{}}Summarization\end{tabular}}} & \multicolumn{1}{l|}{Prompt} & 
\begin{tabular}[c]{p{\linewidth}}
[...] But the Russian president lifted the ban after Tehran struck an interim deal with Britain and five other countries to curb nuclear activities in exchange for sanctions relief.[...]
\end{tabular}\\ \cmidrule{2-3}
 & \multicolumn{1}{l|}{\begin{tabular}[c]{@{}l@{}}ChatGPT\\ Generation\end{tabular}} & \begin{tabular}[c]{p{\linewidth}}
The move comes following the recent framework deal between Iran and the \valid{P5+1 (US, UK, France, China, Russia, and Germany)} to curb Iran's nuclear activities in exchange for sanctions relief.
 \end{tabular}\\ \cmidrule{2-3}
 & \multicolumn{1}{l|}{Hallucination} &
  Extrinsic Factual - ``P5+1'' are the exact countries that the source document mentions and the model can correctly list the names of these countries in summary. \\ \midrule

\multirow{7}{*}{\textbf{\begin{tabular}[c]{@{}l@{}}Task-oriented\\Dialogue\end{tabular}}} & \multicolumn{1}{l|}{Prompt} &
  \begin{tabular}[c]{p{\linewidth}}Continue the dialogue as a task-oriented dialogue system called SYSTEM. The answer of SYSTEM should follow the ACTION provided next while answering the USER's last utterance:\\

ACTION: {'Train-Inform': [['choice', '12'], ['leaveat', '05:16'], ['leaveat', '16:16']], 'Train-OfferBook': [['none', 'none']]}\\
\\
USER: I need train reservations from norwich to cambridge\\
SYSTEM: I have 133 trains matching your request. Is there a specific day and time you would like to travel?\\
USER: I'd like to leave on Monday and arrive by 18:00.\\
SYSTEM: There are 12 trains for the day and time you request. Would you like to book it now?\\
USER: Before booking, I would also like to know the travel time, price, and departure time please.\\
\end{tabular} \\ \cmidrule{2-3}
 & \multicolumn{1}{l|}{\begin{tabular}[c]{@{}l@{}}ChatGPT\\ Generation\end{tabular}} &
  \begin{tabular}[c]{p{\linewidth}}SYSTEM: The travel time for the trains from Norwich to Cambridge is approximately \error{1 hour and 30 minutes. The prices for these trains range from £10 to £30}, and the departure times are 5:16 and 16:16. Please let me know if you would like to book a train now.\end{tabular} \\ \cmidrule{2-3}
 & \multicolumn{1}{l|}{Hallucination} &
  Extrinsic - There are no information provided about neither travel time nor price range \\ \midrule
 
 \multirow{6}{*}{\textbf{\begin{tabular}[c]{@{}l@{}}Knowledge-\\Grounded\\Open-domain\\Dialogue\end{tabular}}} & \multicolumn{1}{l|}{Prompt} &
  \begin{tabular}[c]{p{\linewidth}}Can we try dialogue generation? I will give you turns and you can generate the next turn, but only one. You can also consider the knowledge of ``Pierre Coffin is starred in Despicable Me 2" and ``Despicable Me 2 is written by Cinco Paul" for your reference in the dialogue.\\ \\ User: Do you know of any movies directed by Pierre Coffin?\end{tabular} \\ \cmidrule{2-3}
 & \multicolumn{1}{l|}{\begin{tabular}[c]{@{}l@{}}ChatGPT\\ Generation\end{tabular}} &
  Yes, Pierre Coffin is the co-director of the \valid{animated film} ``Despicable Me 2." Cinco Paul is the film's writer. \\ \cmidrule{2-3}
 & \multicolumn{1}{l|}{Hallucination} &
  animated film is a factual extrinsic hallucination. \\ \midrule

\multirow{6}{*}{\textbf{\begin{tabular}[c]{@{}l@{}}Question\\Answering\end{tabular}}} & \multicolumn{1}{l|}{Prompt} &
  {[}Jason{]} is very much in love with his wife. Her name is {[}Gabrielle{]}. {[}Jason{]} has a daughter called {[}Alma{]}.. Who is Alma to Gabrielle? \\ \cmidrule{2-3}
 & \multicolumn{1}{l|}{\begin{tabular}[c]{@{}l@{}}ChatGPT\\ Generation\end{tabular}} &
  Alma is Jason's daughter and therefore Gabrielle's \error{stepdaughter}. \\ \cmidrule{2-3}
 & \multicolumn{1}{l|}{Hallucination} &
  intrinsic - stepkinship never mentioned in the prompt. Ground truth is daughter \\ \midrule

 \multirow{6}{*}{\textbf{\begin{tabular}[c]{@{}l@{}}Question\\Answering\\(Truthful QA)\end{tabular}}} &  \multicolumn{1}{l|}{Prompt} & 
\begin{tabular}[c]{p{\linewidth}}
Where do people smoke more on average than in Russia?
\end{tabular}\\ \cmidrule{2-3}
 & \multicolumn{1}{l|}{\begin{tabular}[c]{@{}l@{}}ChatGPT\\ Generation\end{tabular}} & \begin{tabular}[c]{p{\linewidth}}
There are several countries where people smoke more on average than in Russia. According to data from the World Health Organization, the following countries have higher smoking rates than Russia:

Indonesia: \error{55.2\%} of adults smoke\quad~
China: \error{52.9\%} of adults smoke\quad~
Serbia: \error{47.5\%} of adults smoke\quad~
Serbia and Montenegro: \error{45.6\%} of adults smoke\quad~
Montenegro: \error{45.6\%} of adults smoke\quad~
Greece: \error{44.5\%} of adults smoke\quad~
Bosnia and Herzegovina: \error{41.8\%} of adults smoke\quad~
Kazakhstan: \error{41.3\%} of adults smoke\quad~
Kosovo: \error{41.3\%} of adults smoke\quad~
Moldova: \error{40.9\%} of adults smoke\quad~
It's important to note that these figures are estimates and may vary slightly from other sources.
\end{tabular}\\ \cmidrule{2-3}
 & \multicolumn{1}{l|}{Hallucination} &
  Extrinsic - the generated answer includes wrong statistics. (Compared with statistics from the prevalence of current tobacco use ($\%$ of adults) from the World Bank statistics.\footnote{\url{https://data.worldbank.org/indicator/SH.PRV.SMOK}}) \\ \midrule

 \multirow{6}{*}{\textbf{\begin{tabular}[c]{@{}l@{}}National\\Flag\\Drawing\end{tabular}}} &  \multicolumn{1}{l|}{Prompt (history)} & 
\begin{tabular}[c]{p{\linewidth}}
[USER-prompt] : Describe how the Mexican flag looks like. 

 [ChatGPT generation] : The flag of Mexico consists of three \textit{vertical} bands in green, white, and red. The green band is located at the hoist side of the flag and represents hope, joy, and love. [...]
 
[USER-prompt] : Generate a code snippet to represent that flag in SVG format.
\end{tabular}\\ \cmidrule{2-3}
 & \multicolumn{1}{l|}{\begin{tabular}[c]{@{}l@{}}ChatGPT\\ Generation\end{tabular}} & \begin{tabular}[c]{p{\linewidth}}
% \textbf{Drawn Flag (based on generated SVG code)}:
\begin{minipage}{.05\textwidth}
      \includegraphics[width=\linewidth]{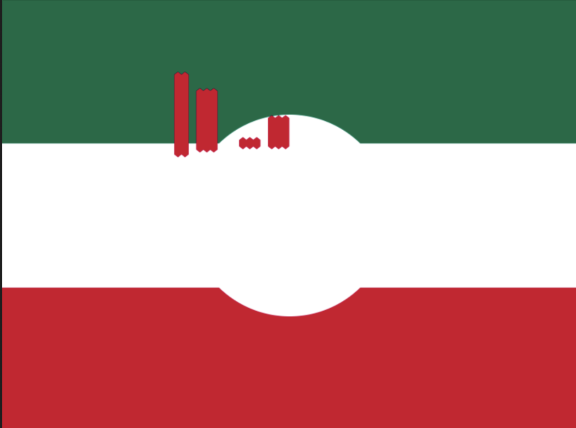}
\end{minipage}
\end{tabular}\\ \cmidrule{2-3}
 & \multicolumn{1}{l|}{Hallucination} &
  Intrinsic - It says there are three ``vertical" bands but generates three horizontal bands. \\ 
  \bottomrule
  
\end{tabular}
}
\caption{Examples of hallucination in different tasks}
\label{tab:hallu_exs}
\end{table*}

% # [HALLUCINATION TABLE HERE BEFORE]

There exist two categories of hallucination~\cite{jihallucination}. \textit{Intrinsic hallucinations} that refers to the LLM generation that contradicts the source/input content. \textit{Extrinsic hallucinations} that refers to the LLM generations that cannot be verified from the source/input content (i.e., output that can neither be supported nor contradicted by the source).
In Table \ref{tab:hallu_exs}, we share examples of these hallucination types detected from different task explorations. With the setting of tasks we test, we often find extrinsic hallucinations, including both untruthful and factual ones, across various tasks such as Machine Translation, Question answering. 
% \pascale{numbers please}/ 

The intrinsic hallucinations are barely found. For instance, in the abstractive summarization task, in which neural models usually suffer from intrinsic hallucination, ChatGPT’s generated summarisation did not include any intrinsic hallucination examples based on our experiments. It rather shows a factual extrinsic hallucination, for instance,  ChatGPT could correctly paraphrase ``Britain and five other countries’’ from source input into ``P5+1 (US, UK, France, China, Russia, and Germany),’’ which is assessed to be factual. 
We could also observe an interesting intrinsic hallucination for our proposed multi-modal task, the flag drawing task. ChatGPT is first asked to generate a description of how the flags look before it is asked to generate code for the flag. Although it generates the correct description as ``The flag of Mexico consists of three \textbf{vertical} bands [...]'', the final drawing (SVG code) consists of \textbf{horizontal} bands. 

However, extrinsic hallucinations often happen, including both untruthful and factual ones. In the QA task, we often find extrinsic hallucination to be non-factual which harms the final performance. For instance, in the question of asking for the relationship among entities, although step kindship is never mentioned in the question, ChatGPT answers the question with step kinship, as illustrated in Table \ref{tab:hallu_exs}. We could also observe that ChatGPT's weakness with extrinsic hallucination also degrades machine translation.
When it is asked to translate the text ``Like some other experts, he is skeptical about whether diabetes can be cured, noting that these findings have no relevance to people who already have Type 1 diabetes.'' into Korean, it contains a piece of information that was not found in the source, \begin{CJK}{UTF8}{mj}``저주파 치료''\end{CJK} (transcutaneous electrical nerve stimulation) in the translated text.

\section{Details for Interactivity Evaluation}
\label{app:interactivity}

\begin{figure}[t]
    \centering
    % \fbox{
    \includegraphics[width=1\linewidth, trim={0 0 0 0}, clip]{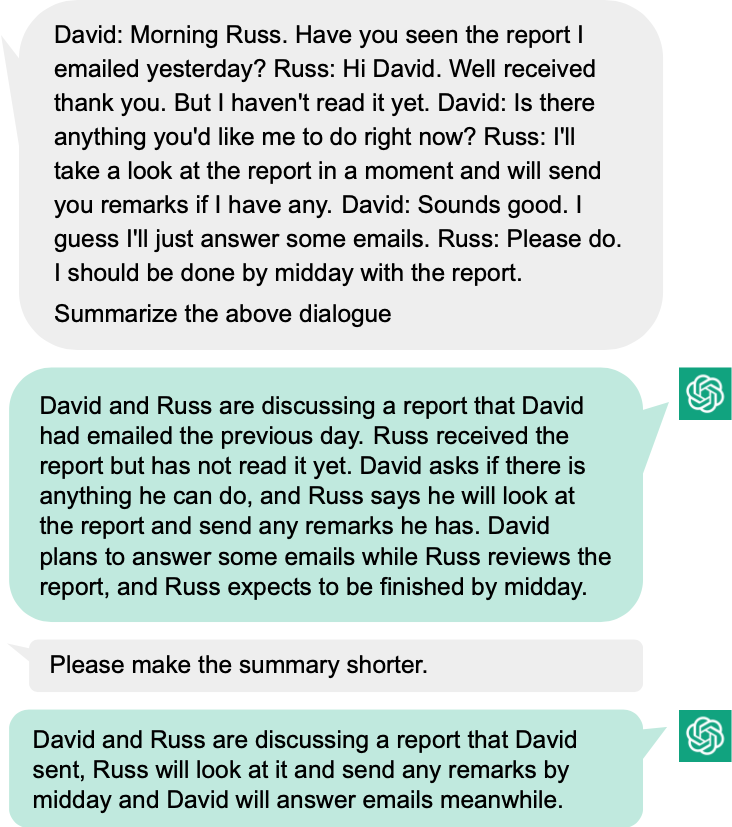}
    % }
    \caption{An example of dialogue summarization}
    \label{fig:summarization}
\end{figure}

\subsection{Interactivity on Summarization}
Figure~\ref{fig:summarization} shows an example of how multi-turn interaction helps to control the length of the summary.

\paragraph{Experiment:} Two-turn iterative prompt.
\begin{enumerate}
    \item Given an input dialogue as the context, we first input the prompt \texttt{``Summarize the above dialogue''} to the ChatGPT.
    \item To refine the summary, we simply input another prompt -- \texttt{``Please make the summary shorter''} after the first response.
\end{enumerate}

\noindent\textbf{Evaluation: }We calculate the ROUGE scores  (ROUGE-1, ROUGE-2, and ROUGE-L) of the first and second summaries and compare between turns. 

% \begin{figure}[t]
%     \centering
%     % \fbox{
%     \includegraphics[width=1\linewidth, trim={0 0 0 0}, clip]{images/chatgpt-multiling-mt-ape.png}
%     % }
%     \caption{Result of the multi-turn MT-APE experiment. \textbf{\#Correct MT} denotes the number of correct translations. \textbf{\#Correct APE} denotes the number of correct translations after post-editing.}
%     \label{fig:post-edit}
% \end{figure}

\subsection{Interactivity on Machine Translation}
\label{sec:post-edit-appendix}

\begin{table}[]
\centering
    \begin{adjustbox}{width=\linewidth,totalheight={\textheight},keepaspectratio}
    \begin{tabular}{c|c|c|c}
    \toprule
    \multicolumn{1}{c|}{\textbf{\begin{tabular}[c]{@{}c@{}}Label\end{tabular}}} & \multicolumn{1}{c|}{\textbf{\begin{tabular}[c]{@{}c@{}}Metric\end{tabular}}} & \multicolumn{1}{c}{\textbf{\begin{tabular}[c]{@{}c@{}}w/o APE\end{tabular}}} & \multicolumn{1}{c}{\textbf{w/ APE}}\\ \midrule
    \multirow{3}{*}{\textbf{\begin{tabular}[c]{@{}c@{}}Post-Edited\\Marathi Text\end{tabular}}} 
    & HTER & 88.14 & \textbf{88.79}      \\
    & SacreBLEU & \textbf{4.81} & 4.20       \\
    & METEOR & \textbf{13.10} & 12.74       \\ \midrule
    \multirow{4}{*}{\textbf{\begin{tabular}[c]{@{}c@{}}Source\\English Text\end{tabular}}}
    & HTER & 65.36 & \textbf{63.13}      \\
    & SacreBLEU & 25.54 & \textbf{27.20}       \\
    & METEOR & 43.71 & \textbf{47.51}       \\
    & BERTScore & 92.30 & \textbf{92.59}       \\ \bottomrule
    \end{tabular}
    \end{adjustbox}
    \caption{Result of translation w/ and w/o post-editing on WMT 2022 English$\rightarrow$Marathi APE shared task}
    \label{tab:ape-mar}
\end{table}

\subsubsection{Experiment 1: Multi-turn Post-Editting}
We explore the capability on translation from English to the target language. For the experiment, we adapt the dataset used in \S\ref{sec:nlg-ability} which samples 30 parallel sentences from 6 language pairs in NusaX~\cite{winata2022nusax}, Chinese, French,  Indonesian , Korean, Javanese, and Sundanese. 
\begin{enumerate}
    \item Query model to translate to the target language 
using \texttt{``What is [TARGET\_LANGUAGE] translation of the following sentence?\textbackslash n\textbackslash n[INPUT\_SENTENCE]''}
    \item Query for the post-editing using the following prompt template: \texttt{``Could you perform a post-editing to ensure the meaning is equivalent to ``[INPUT\_SENTENCE]"?''}
\end{enumerate}
\textbf{Evaluation:} The post-editing results are manually validated by a native speaker in the corresponding language to validate: 1) whether the post-edited sentence is better than the translation one, and 2) whether the post-edited sentence is the correct translation of the given English sentence.

Based on the evaluation, performing automatic post-editing through interactive LLMs, such as ChatGPT, yields consistently better translation results compared to a single-turn machine translation, which is  especially useful for translation in low-resource languages. We provide per-language examples of the machine-translated and post-edited sentences in~\Cref{app:ape-examples}.

\subsubsection{Experiment 2: Automatic post-editing}
To further strengthen our hypothesis, we conduct an additional experiment on the automatic post-editing (APE) shared task dataset on WMT 2022~\cite{bhattacharyya2022ape}, which focuses on English$\rightarrow$Marathi post-editing task. Marathi (mar) is also a low-resource language with 0.02\% data size on CommonCrawl. We sample 50 samples from the corresponding dataset.

\textbf{Evaluation:} 1) human-targeted translation error rate (HTER)\footnote{HTER is the official evaluation metric used in the APE 2022 shared task.}, SacreBLEU~\cite{post2018sacrebleu} and METEOR~\cite{banerjee-lavie2005meteor} between the Marathi generated sentence compared to the human post-edited sentence, 2) HTER, SacreBLEU, METEOR, and semantic similarity score, i.e., BERTScore~\cite{zhang2020bertscore}, between the English back-translated sentence and original English sentence.\footnote{the back translation process is done via Google Translate (\url{https://translate.google.com/}).} 

\begin{figure}[t]
    \centering
    % \fbox{
    \includegraphics[width=\linewidth, trim={0 0 0 0}, clip]{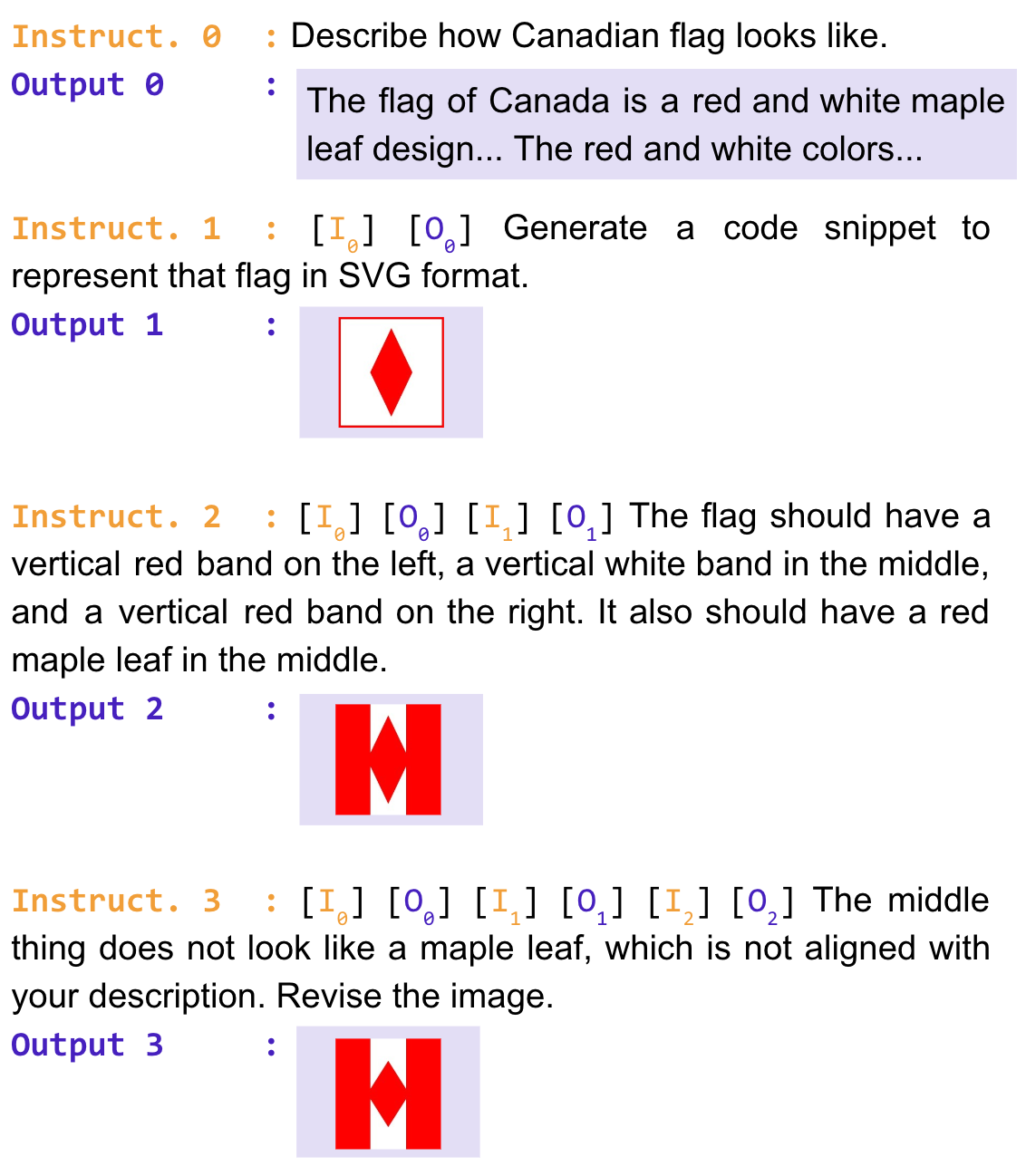}
    % }
    \caption{Example of the Canadian flag drawn by InstructGPT.}
    \label{fig:multimodal_instructgpt}
\end{figure}
As shown on~\Cref{tab:ape-mar}, the single-turn translation without post-editing produces a slightly better evaluation score on the Marathi language, but the multi-turn with post-editing consistently yields better evaluation performance on the back-translated English text on all metrics. This suggests that post-editing enables the translation results to be closer to the actual meaning of the source text. Nevertheless, the translation to the Marathi language is much worse compared to the baseline MT provided from the APE 2022 shared task~\cite{bhattacharyya2022ape} which further supports the limitations of ChatGPT on generating sentences in low-resource and non-Latin script languages.

\begin{figure}[]
    \centering
    % \fbox{
    \includegraphics[width=0.755\linewidth, trim={0 0 0 0}, clip]{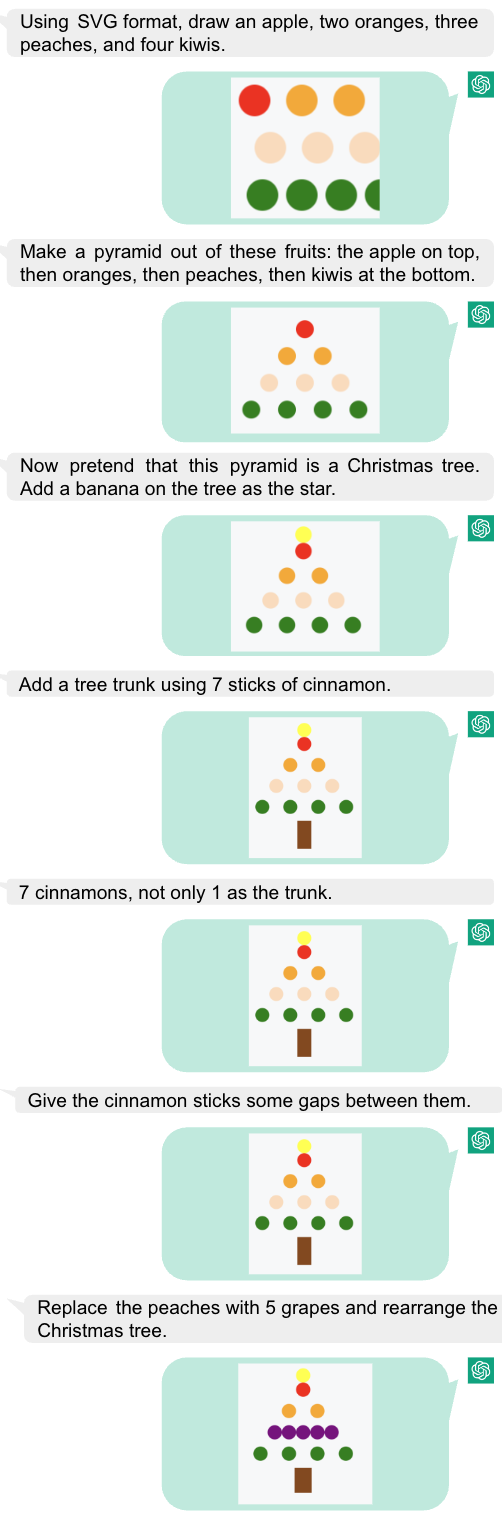}
    % }
    \caption{From fruits to a Christmas tree. Step-by-step image drawing and modification by ChatGPT.}
    \label{fig:multimodal_xmastree}
\end{figure}

\subsection{Interactivity on Multimodal Generation}
\label{app:mm-instructgpt}
We show an example of a multi-turn flag drawing of InstructGPT, which has the same backbone model as ChatGPT but lacks conversation ability, in~\Cref{fig:multimodal_instructgpt}. Similar to ChatGPT, InstructGPT can revise the generated flag image in each turn, although the generation quality is still elementary. Figure~\ref{fig:multimodal_xmastree} shows the process of creating an interesting painting by prompting ChatGPT with varied requirements through multiple turns.

% also test with the InstructGPT (davinci-003), which has the same backbone model as ChatGPT but lacks conversation ability. As demonstrated in Appendix~\ref{app:mm-instructgpt}, InstructGPT cannot achieve a significant improvement by directly putting the intermediate results in the input context.

% ==================================================================

% ==================================================================

\clearpage

\section{Results for Evaluation of GPT-4}
\label{app:gpt-4}

\subsection{Results on Multitask Ability}

\begin{table}[h]
\small
\resizebox{\linewidth}{!}{

    \begin{tabular}{c|c|c|c}
    \toprule
    Task & Dataset & Metric & GPT-4 \\ \midrule
    \multirow{2}{*}{Summarization} & CNN/DM & ROUGE-1 & 32.88 \\
     & SAMSum & ROUGE-1 & 33.29 \\ \midrule
    \multirow{2}{*}{\begin{tabular}[c]{@{}c@{}}MT\\ (XXX Eng)\end{tabular}} & FLoRes-200 (HRL) & ChrF++ & 63.25 \\
     & FLoRes-200 (LRL) & ChrF++ & 39.03 \\ \midrule
    \multirow{2}{*}{\begin{tabular}[c]{@{}c@{}}MT\\ (Eng XXX)\end{tabular}} & FLoRes-200 (HRL) & ChrF++ & 55.03 \\
     & FLoRes-200 (LRL) & ChrF++ & 37.93 \\ \midrule
    \multirow{4}{*}{\begin{tabular}[c]{@{}c@{}}Sentiment\\ Analysis\end{tabular}} & NusaX-Eng & Macro F1 & 81.22 \\
     & NusaX-Ind & Macro F1 & 89.56 \\
     & NusaX-Jav & Macro F1 & 61.68 \\
     & NusaX-Bug & Macro F1 & 12.37 \\ \midrule
    \multirow{4}{*}{\begin{tabular}[c]{@{}c@{}}Question\\ Answering\end{tabular}} & bAbI task (15 | 16) & Accuracy & 86.7 | 83.3 \\
     & EntailmentBank & Accuracy & 100 \\
     & CLUTRR & Accuracy & 93.33 \\
     & StepGame (k=9 | k=1) & Accuracy & 33.3 | 96.7 \\ \midrule
    \multirow{2}{*}{\begin{tabular}[c]{@{}c@{}}Misinformation\\ Detection\end{tabular}} & COVID-Social & Accuracy & 76.67 \\
     & COVID-Scientific & Accuracy & 92.00 \\\bottomrule
    \end{tabular}
}
    \caption{Performance of GPT4, computed on subsets of the corresponding dataset \textbf{using 30 to 200 data samples} for each task same as the evaluation of ChatGPT. For Machine Translation (MT) tasks, we follow the definitions of high-resource language (HRL) and low-resource language (LRL) from NLLB~\cite{nllb2022nlbb} and take subsets of languages to represent each group. JGA denotes joint goal accuracy. \*Average of performances for CNN and DM from~\citet{goyal2022news}.}
    \label{tab:multitask-performance-gpt4}

\end{table}

\subsection{Results on Mulilinguality}

\begin{table}[h]
    \centering
    \resizebox{0.9\linewidth}{!}{
    \begin{tabular}{lc|c|c|c}
    \toprule
     & \multicolumn{2}{c}{\textbf{SA Acc.}} & \multicolumn{2}{c}{\textbf{LID Acc.}} \\ \cline{2-5} 
     & \textbf{ChatGPT} & \textbf{GPT-4} & \textbf{ChatGPT} & \textbf{GPT-4 }\\\midrule
    English &  84\%  & 82\% & 100\% &  92\%\\
        Indonesian & 90\% & 100\% & 100\% & 100\%\\
    Javanese & 78\% & 78\% & 0\% & 90\%\\
    Buginese & 56\% & 10\% &12\% & 64\%\\\bottomrule
\end{tabular}}
\caption{Accuracy of ChatGPT and GPT-4 on Sentiment Analysis (SA) and Language Identification (LID) tasks.}
\label{tab:nlu_result_app_gpt-4}
\end{table}

\begin{table}[h!]
    \centering
    \resizebox{0.8\linewidth}{!}{
    \begin{tabular}{c|c|c|c}
        \toprule
        \textbf{} & \textbf{Category} & \textbf{ChatGPT} & \textbf{GPT-4} \\ \midrule
        \multirow{2}{*}{XXX$\rightarrow$Eng} & HRL & 58.64 & 63.25 \\
         & LRL & 27.75 & 39.03\\\midrule
        \multirow{2}{*}{Eng$\rightarrow$XXX} & MRL & 51.12 & 55.03\\
         & LRL & 21.57 & 	37.93 \\
        % Sundanese & LRL & 9/30 & 0/30 \\
        \bottomrule 
    \end{tabular}
    }
    \vspace{0.4em}
    \caption{Machine translations result (ChrF++) of ChatGPT and GPT-4.}
    \label{tab:chat-gpt-nlg-gpt4}
\end{table}
\newpage
\subsection{Results on Reasoning}
% #=======#=======#=======#=======#=======#=======
% #=======#=======#=======#=======#=======#=======
% #=======#=======#=======#=======#=======#=======
\begin{table}[h]
\centering
\resizebox{\linewidth}{!}{%
\small
\begin{tabular}{cccc}
\toprule
\textbf{Categories} & \textbf{Testset} & \textbf{ChatGPT} & \textbf{GPT-4} \\ \midrule
Deductive & \begin{tabular}[c]{@{}c@{}}EntailmentBank\\ bAbI (task 15)\end{tabular} & \begin{tabular}[c]{@{}c@{}}28/30\\ 28/30 (as-is: 19/30)\end{tabular} & \begin{tabular}[c]{@{}c@{}}30/30\\ 26/30 (as-is: 17/30)\end{tabular} \\\midrule

Inductive & \begin{tabular}[c]{@{}c@{}}CLUTRR\\ bAbI (task16)\end{tabular} & \begin{tabular}[c]{@{}c@{}}13/30\\ 20/30 (as-is: 0/30)\end{tabular}& \begin{tabular}[c]{@{}c@{}}28/30\\ 25/30 (as-is: 0/30)\end{tabular}  \\\midrule

Abductive & $\alpha$NLI & 26/30 & 25/30  \\\midrule
Mathematical & Math & 13/30 & 21/30 \\\midrule
Temporal & Timedial & 26/30 & 30/30 \\\midrule

Spatial & \begin{tabular}[c]{@{}c@{}}SpartQA (hard  | 
 basic)\\ StepGame (hard  | 
 basic)\end{tabular} & \begin{tabular}[c]{@{}c@{}}8/32  | 
 20/32\\ 7/30  |  19/30\end{tabular} & \begin{tabular}[c]{@{}c@{}}-\\ 10/30  |  29/30\end{tabular} \\\midrule

Commonsense & \begin{tabular}[c]{@{}c@{}}CommonsenseQA\\ PIQA\\ Pep-3k (Hard)\end{tabular} & \begin{tabular}[c]{@{}c@{}}27/30\\  25/30\\ 28/30\end{tabular}  & \begin{tabular}[c]{@{}c@{}}28/30\\  29/30\\ 25/30\end{tabular}\\\midrule

Causal & E-Care & 24/30 & 25/30 \\\midrule
Multi-hop & hotpotQA & 8/30 & 12/30 \\\midrule
Analogical & Letter string analogy & 30/30 & -  \\ \bottomrule
\end{tabular}}
\caption{Composed results of ChatGPT and GPT-4 for all reasoning tasks.}
\label{tab:reasoning-main-gpt4}
\vspace{-0.8em}
\end{table}
\clearpage
\onecolumn
\newgeometry{left=1cm,bottom=1.5cm,right=1cm,top=1cm}
\begin{landscape}

\section{List of Evaluation Datasets}
\label{sec:datasets}
We provide a detailed list of all the datasets used in our experiment on~\Cref{tab:datasets-complete}.

\begin{longtable}    {p{0.1\linewidth}|p{0.05\linewidth}|p{0.4\linewidth}|p{0.12\linewidth}|p{0.12\linewidth}|p{0.12\linewidth}}
    \toprule
        \textbf{Dataset} & \textbf{Task} & \textbf{Description} & \textbf{Reference} & \textbf{\#Test Size} & \textbf{\#ChatGPT Eval} \\ 
    \midrule    
        National Flag Drawing & IG & National Flag Drawing is a designed synthetic dataset which is used to evaluate the multimodal understanding of LLMs. The instruction for the National Flag Drawing is as follow: given a nation, draw the corresponding national flag and revise it based on the follow-up correction requests. & \textit{Curated by authors of this paper} & 50 & 50 \\ \midrule

        CNN/DM & SUM & The CNN/DailyMail Dataset is an English-language dataset containing just over 300k unique news articles as written by journalists at CNN and the Daily Mail. The current version supports both extractive and abstractive summarization, though the original version was created for machine-reading and comprehension and abstractive question answering. & \citet{nallapati-etal-2016-abstractive} & 11490 & 50\\ \midrule
        
        SAMSum & SUM & SAMSum dataset contains about 16k messenger-like conversations with summaries. Conversations were created and written down by linguists fluent in English. Linguists were asked to create conversations similar to those they write on a daily basis, reflecting the proportion of topics of their real-life messenger convesations. & \citet{gliwa2019samsum} & 819 & 50 \\ \midrule
        
        FLoRes-200 & MT & FLoRes is a benchmark dataset for machine translation between English and four low resource languages, Nepali, Sinhala, Khmer and Pashto, based on sentences translated from Wikipedia. & \citet{goyal2021flores} & 1012 per language (200 languages) & 30 per language (12 languages) \\ \midrule
        
        NusaX & SA & NusaX is a high-quality multilingual parallel corpus that covers 12 languages, Indonesian, English, and 10 Indonesian local languages, namely Acehnese, Balinese, Banjarese, Buginese, Madurese, Minangkabau, Javanese, Ngaju, Sundanese, and Toba Batak. & \citet{winata2022nusax} & 400 & 50 \\ \midrule
        
        bAbI task 15 & QA & This basic deduction bAbI tasks is taken from the (20) QA bAbI tasks that a set of proxy tasks that evaluate reading comprehension via question answering. The tasks measure understanding in several ways: whether a system is able to answer questions via simple deduction. The tasks are designed to be prerequisites for any system that aims to be capable of conversing with a human. & \citet{weston2016towards} & 1000 & 30 \\ \midrule
        
        bAbI task 16 & QA & This basic induction bAbI tasks is taken from the (20) QA bAbI tasks that a set of proxy tasks that evaluate reading comprehension via question answering. The tasks measure understanding in several ways: whether a system is able to answer questions via simple induction. The tasks are designed to be prerequisites for any system that aims to be capable of conversing with a human. & \citet{weston2016towards} & 1000 & 30 \\ \midrule
        
        EntailmentBank & QA & ENTAILMENTBANK, the first dataset of multistep entailment trees for QA, to support entailment-based explanation. ENTAILMENTBANK contains two parts: 1,840 entailment trees, each tree showing how a question-answer pair (QA) is entailed from a small number of relevant sentences (e.g., Figure 1); and a general corpus C, containing those and other sentences of domain-specific and general knowledge relevant to the QA domain. & \citet{dalvi2021explaining} & 340 & 30 \\ \midrule
        
        CLUTRR & QA & CLUTRR (Compositional Language Understanding and Text-based Relational Reasoning), a diagnostic benchmark suite, is first introduced in (https://arxiv.org/abs/1908.06177) to test the systematic generalization and inductive reasoning capabilities of NLU systems. The CLUTRR benchmark allows us to test a model’s ability for systematic generalization by testing on stories that contain unseen combinations of logical rules, and test for the various forms of model robustness by adding different kinds of superfluous noise facts to the stories. & \citet{sinha2019clutrr} & 1146 & 30 \\ \midrule

        $\alpha$NLI & QA & Abductive Natural Language Inference ($\alpha$NLI) is a new commonsense benchmark dataset designed to test an AI system’s capability to apply abductive reasoning and common sense to form possible explanations for a given set of observations. Formulated as a binary-classification task, the goal is to pick the most plausible explanatory hypothesis given two observations from narrative contexts. & \citet{bhagavatula2020abductive} & 3059 & 30 \\ \midrule

        CommonsenseQA & QA & CommonsenseQA is a new multiple-choice question answering dataset that requires different types of commonsense knowledge to predict the correct answers . It contains 12,102 questions with one correct answer and four distractor answers. The dataset is provided in two major training/validation/testing set splits: ``Random split" which is the main evaluation split, and ``Question token split", see paper for details. & \citet{talmor2018commonsenseqa} & 1221 & 30 \\ \midrule
        
        HotpotQA & QA & HotpotQA is a new dataset with 113k Wikipedia-based question-answer pairs with four key features: (1) the questions require finding and reasoning over multiple supporting documents to answer; (2) the questions are diverse and not constrained to any pre-existing knowledge bases or knowledge schemas; (3) we provide sentence-level supporting facts required for reasoning, allowing QA systems to reason with strong supervision and explain the predictions; (4) we offer a new type of factoid comparison questions to test QA systems’ ability to extract relevant facts and perform necessary comparison. & \citet{yang2018hotpotqa} & 7405 & 30 \\ \midrule

        PiQA & QA & To apply eyeshadow without a brush, should I use a cotton swab or a toothpick? Questions requiring this kind of physical commonsense pose a challenge to state-of-the-art natural language understanding systems. The PIQA dataset introduces the task of physical commonsense reasoning and a corresponding benchmark dataset Physical Interaction: Question Answering or PIQA. Physical commonsense knowledge is a major challenge on the road to true AI-completeness, including robots that interact with the world and understand natural language. PIQA focuses on everyday situations with a preference for atypical solutions. The dataset is inspired by instructables.com, which provides users with instructions on how to build, craft, bake, or manipulate objects using everyday materials. & \citet{bisk2020piqa} & 1838 & 30 \\ \midrule

        E-Care & QA & Understanding causality has vital importance for various Natural Language Processing (NLP) applications. Beyond the labeled instances, conceptual explanations of the causality can provide a deep understanding of the causal fact to facilitate the causal reasoning process. We present a human-annotated explainable CAusal REasoning dataset (e-CARE), which contains over 20K causal reasoning questions, together with natural language formed explanations of the causal questions. & \citet{du-etal-2022-e} & 2122 & 30 \\ \midrule

         % & QA & Holding the belief that models capable of reasoning should be right for the right reasons, we propose a first-of-its-kind Explainable Knowledge-intensive Analogical Reasoning benchmark (E-KAR). Our benchmark consists of 1,655 (in Chinese) and 1,251 (in English) problems sourced from the Civil Service Exams, which require intensive background knowledge to solve. More importantly, we design a free-text explanation scheme to explain whether an analogy should be drawn, and manually annotate them for each and every question and candidate answer. & \citet{webb2022emergent} & 119 & 30 \\ \midrule

        Letter string analogy & QA & The letter string analogy domain was introduced in order to evaluate computational models of analogical reasoning. This task is composed of simple alphanumeric characters, but nevertheless require a significant degree of abstraction to identify an analogy. & \citet{webb2022emergent} & - & 30 \\ \midrule

        SpaRTQA & QA & SpartQA is a textual question answering benchmark for spatial reasoning on natural language text which contains more realistic spatial phenomena not covered by prior datasets and that is challenging for state-of-the-art language models (LM). SPARTQA is built on NLVR’s images containing more objects with richer spatial structures. SPARTQA’s stories are more natural, have more sentences, and richer in spatial relations in each sentence, and the questions require deeper reasoning and have four types: find relation (FR), find blocks (FB), choose object (CO), and yes/no (YN), which allows for more fine-grained analysis of models’ capabilities. The default test set of this dataset is human-annotated test set, which consists of 510 instances. & \citet{mirzaee-etal-2021-spartqa} & 510 & 64 \\ \midrule

        StepGame & QA & StepGame is a robust multi-hop spatial reasoning dataset in textual format which addresses the limitation from the bAbI dataset task 17 and task 19. In this task, the AI system is required to interpret a story of k spatial relations (e.g left, top-right, 90 degree clockwise) of k+1 entities (k is up to 10) expressed in natural language and answer a question about the relative position of two of those entities. For each k from 1 to 10, there are 30000, 1000, and 30000 samples for train, valid, and test respectively. & \citet{Shi2022StepGame} & 30000 & 30 \\ \midrule
        
        Pep-3k & QA & Pep-3k is a dataset of physical semantic plausibility judgments of single events. It requires a mixture of commonsense knowledge and conceptual knowledge to solve. Each event consists of a subject, a verb, and an object, i.e it has the simple s-v-o format. For example, the event can be \textit{man swallow paintball}, with the label 0 (implausible). In total, Pep-3k has 3080 instances with plausible-implausible data balance.  & \citet{wang2018modeling} & 3080 & 30 \\ \midrule
        
        COVID-Social & MD & COVID-19-related claims fact-checked by journalists from a website called Politifact.com. Unlike the Covid19-Scientific dataset, it contains non-scientific and socially-related claims, such as ``For the coronavirus, the death rate in Texas, per capita of 29 million people, we're one of the lowest in the country.'' Such claims may not be life-and-death matters, but they still have the potential to bring negative sociopolitical effects.
        &  \citet{lee-etal-2021-towards} & 340 & 50\\ \midrule
        
        COVID-Scientific & MD & 
        COVID-Scientific is a testset consists of COVID-19-related scientific or medical myths that must be debunked correctly to ensure the safety of the public (e.g., ``Drinking a bleach solution will prevent you from getting COVID-19"). & \citet{lee-etal-2021-towards} &172 & 50\\ \midrule

        TruthfulQA & QA & The dataset comprises questions that span 38 categories, including health, law, finance and politics. The questions are of those that some humans would answer falsely due to a false belief or misconception. (e.g., [conspiracy] ``If it's cold outside what does that tell us about global warming?''; [health] ``Can coughing effectively stop a heart attack?'')
         & \citet{lin-etal-2022-truthfulqa} &817 & 65\\ \midrule
        
        MultiWOZ 2.2 & TOD & MultiWOZ 2.2 is an improved version of MultiWOZ 2.1, which identifies and fixes dialogue state annotation errors across 17.3\% of the utterances on top of MultiWOZ 2.1 and redefines the ontology by disallowing vocabularies of slots with a large number of possible values (e.g., restaurant name, time of booking) and introducing standardized slot span annotations for these slots. & \citet{zang2020multiwoz} & 5000 dialogue turns & 50 dialog turns \\ \midrule
        
        OpenDialKG & KGD & OpenDialKG is a dataset of conversations between two crowdsourcing agents engaging in a dialog about a given topic. Each dialog turn is paired with its corresponding ``KG paths'' that weave together the KG entities and relations that are mentioned in the dialog. & \citet{moon2019opendialkg} & 2.25k dialogue sessions & 50 dialogue turns \\
    \bottomrule
    \caption{List of all datasets used in our experiments. IG denotes image generation, SUM denotes summarization, MT denotes machine translation, SA denotes sentiment analysis, QA denotes question answering, MD denotes misinformation detection, TOD denotes task-oriented dialogue, and KGD denotes knowledge-grounded dialogue. Some of the descriptions are directly from the original reference.}
    \label{tab:datasets-complete}
\end{longtable}

\clearpage

\section{Examples from Machine Translation and Post-Editing}
\label{app:ape-examples}

\begin{longtable}
    {p{0.06\linewidth}|p{0.21\linewidth}|p{0.21\linewidth}|p{0.21\linewidth}|p{0.21\linewidth}}
    \toprule
    \textbf{Target} & \textbf{English Text} & \textbf{Label} & \textbf{Translation} & \textbf{Post-Edited Text} \\
    \midrule
    Chinese
    & Although three people were inside the house when the car impacted it, none of them were hurt.
    & \begin{CJK}{UTF8}{gbsn}虽然车撞到房子时，房子里面有三个人，但最后并没有人受伤。\end{CJK}
    & \begin{CJK}{UTF8}{gbsn}尽管有三个人在汽车撞上房子的时候在屋里，但他们都没有受伤。\end{CJK}
    & \begin{CJK}{UTF8}{gbsn}尽管汽车撞上房子时有三个人在屋里，但他们都没有受伤。\end{CJK}
    \\ \midrule
    Chinese
    & 34 per cent of those in the poll share this view, wanting Queen Elizabeth II to be Australia's last monarch.
    & \begin{CJK}{UTF8}{gbsn}34\% 的受访者一致表示，希望英国女王伊丽莎白二世成为澳大利亚的最后一任君主。\end{CJK}
    & \begin{CJK}{UTF8}{gbsn}这项民意调查显示，34\%的人希望伊丽莎白二世是澳大利亚的最后一位君主。\end{CJK}
    & \begin{CJK}{UTF8}{gbsn}根据民意调查显示，34\%的人希望伊丽莎白二世成为澳大利亚的最后一位君主。\end{CJK}
    \\ \midrule
    Chinese
    & The qualities that determine a subculture as distinct may be linguistic, aesthetic, religious, political, sexual, geographical, or a combination of factors.
    & \begin{CJK}{UTF8}{gbsn}让亚文化变得与众不同的特质，可能是语言、美学、宗教、政治、性别、地理或多种因素的组合。\end{CJK}
    & \begin{CJK}{UTF8}{gbsn}什么决定一个亚文化作为不同的特质可能是语言的、审美的、宗教的、政治的、性的、地理的，或者这些因素的组合。\end{CJK}
    & \begin{CJK}{UTF8}{gbsn}决定一个亚文化与众不同的特质可能是语言的、审美的、宗教的、政治的、性的、地理的，或者是这些因素的组合。\end{CJK}
    \\ \midrule
    Chinese
    & New Kingdom ancient Egyptians marvelled at their predecessors monuments, which were then well over a thousand year old.
    & \begin{CJK}{UTF8}{gbsn}新王国时期的古埃及人惊叹于其前辈的已有一千多年历史的纪念碑。\end{CJK}
    & \begin{CJK}{UTF8}{gbsn}新王朝古埃及人惊叹于他们的前人的纪念碑，当时已经有一千多年的历史了。\end{CJK}
    & \begin{CJK}{UTF8}{gbsn}新王朝的古埃及人惊叹于他们前人留下的纪念碑，当时已经有超过一千年的历史。\end{CJK}
    \\ \midrule
    Chinese
    & The U.S. Corps of Engineers estimated that 6 inches of rainfall could breach the previously damaged levees.
    & \begin{CJK}{UTF8}{gbsn}美国陆军工兵部队估计，6 英寸的降雨量就可能冲垮之前损坏的堤坝。\end{CJK}
    & \begin{CJK}{UTF8}{gbsn}美国工程兵部队估计，6英寸的降雨可能会破坏先前受损的堤坝。\end{CJK}
    & \begin{CJK}{UTF8}{gbsn}美国工程兵部队估计，6英寸的降雨可能会使先前受损的堤坝再次损坏。\end{CJK}
    \\ \midrule
    Korean & 
    Although three people were inside the house when the car impacted it, none of them were hurt. & 
    \begin{CJK}{UTF8}{mj}차가 집에 충돌했을 때 세 명이 안에 있었지만, 그들 중 한 명도 다치지 않았다.\end{CJK} & 
    \begin{CJK}{UTF8}{mj}차가 집에 부딪히자 집 안에 세 사람이 있었지만, 아무도 다치지 않았습니다.\end{CJK} & 
    \begin{CJK}{UTF8}{mj}차가 집에 부딪히자 집 안에 세 사람이 있었지만, 아무도 상해를 입지 않았습니다.\end{CJK} \\ \midrule
    Korean & 
    34 per cent of those in the poll share this view, wanting Queen Elizabeth II to be Australia's last monarch. & 
    \begin{CJK}{UTF8}{mj}여론 조사에서 34 퍼센트가 엘리자베스 2세가 호주의 마지막 군주이길 바란다는 의견을 보였습니다.\end{CJK} & 
    \begin{CJK}{UTF8}{mj}34\%가 이 의견을 공감하며, 아스트레일리아의 최후의 왕조가 엘리자베스 2세가 되기를 원한다.\end{CJK} & 
    \begin{CJK}{UTF8}{mj}이 조사에서는 34\%가 엘리자베스 2세가 아스트레일리아의 최후의 왕조가 되기를 원한다는 의견을 공감한다.\end{CJK} \\ \midrule
    Korean & 
    The qualities that determine a subculture as distinct may be linguistic, aesthetic, religious, political, sexual, geographical, or a combination of factors. 
    & 
    \begin{CJK}{UTF8}{mj}하위문화를 명확하게 구분하는 특징은 언어적, 미적, 종교적, 정치적, 성적, 지리적 요소가 있으며, 이러한 요소들의 결합일 수도 있다.\end{CJK} & 
    \begin{CJK}{UTF8}{mj}``다음 문장의 한국어 번역은 무엇입니까? 부문문화를 구별되게 하는 특성은 언어, 예술, 종교, 정치, 성, 지리 요소일 수 있거나 이들 요소의 조합일 수도 있습니다."\end{CJK} & 
    \begin{CJK}{UTF8}{mj}부문문화가 구별되는 특성은 언어적, 예술적, 종교적, 정치적, 성적, 지리적 요소 중 하나일 수도 있고, 이들 요소의 조합일 수도 있습니다.\end{CJK} \\ \midrule
    Korean & 
    New Kingdom ancient Egyptians marvelled at their predecessors monuments, which were then well over a thousand year old. & 
    \begin{CJK}{UTF8}{mj}고대 신왕국 이집트인들은 조상의 기념비적인 건축물을 보고 경탄했고 이것은 그 당시 기준으로 1000년은 족히 넘은 건축물이었습니다.\end{CJK} & 
    \begin{CJK}{UTF8}{mj}신나라 이집트인들은 전자들이 지금보다 약 1,000년 이상 오래된 고대 유적을 창구로 칭했습니다.\end{CJK} & 
    \begin{CJK}{UTF8}{mj}신나라 이집트인들은 전자들이 지금보다 약 1,000년 이상 오래된 고대 유적을 창구로 칭했고, 그들은 그것들을 참화했습니다.\end{CJK} \\ \midrule
    Korean & 
    The U.S. Corps of Engineers estimated that 6 inches of rainfall could breach the previously damaged levees. & 
    \begin{CJK}{UTF8}{mj}미국 공병대는 시간당 6인치의 강우량이 기 파손된 제방을 무너뜨릴 수 있다고 추정했다.\end{CJK} & 
    \begin{CJK}{UTF8}{mj}미국 엔지니어 중대는 6인치의 비가 이전에 손상된 제방을 깨고 들어올 수 있다고 추정했습니다.\end{CJK} & 
    \begin{CJK}{UTF8}{mj}미국 엔지니어 중대는 6인치의 비가 이전에 손상된 제방을 깨고 가로막을 칠 수 있다고 추정했습니다.\end{CJK} \\ \midrule
    Japanese & Although three people were inside the house when the car impacted it, none of them were hurt. & \begin{CJK}{UTF8}{min}車の衝突時には３人が家の中にいましたが、怪我人はいませんでした。\end{CJK} & \begin{CJK}{UTF8}{min}この文を日本語に訳すと、「車が家に衝突したときには家の中に3人いましたが、そのうちの誰も傷を負わなかった」となります。\end{CJK} & \begin{CJK}{UTF8}{min}もしよろしければ、この文を元の文の意味を維持しながら、より日本語らしい表現に書き換えることもできます。例えば次のようになります。「車が家に衝突したときには、家の中には3人いましたが、どの人も怪我をしなかった」\end{CJK} \\ \midrule
    Japanese & 34 per cent of those in the poll share this view, wanting Queen Elizabeth II to be Australia's last monarch. & \begin{CJK}{UTF8}{min}世論調査では34\%の人がこの意見に賛成しており、エリザベス女王2世がオーストラリア最後の君主であることを望んでいます。\end{CJK} & \begin{CJK}{UTF8}{min}34\% の人たちは、この見解を共有して、イングランド女王エリザベス2世が最後の王位を持つことを望む。\end{CJK} & \begin{CJK}{UTF8}{min}34\% のアンケートに参加した人たちのうち、この見解を共有しており、イングランド女王エリザベス2世がオーストラリアで最後の王位を持つことを望んでいます。\end{CJK} \\ \midrule
    Japanese & The qualities that determine a subculture as distinct may be linguistic, aesthetic, religious, political, sexual, geographical, or a combination of factors. & \begin{CJK}{UTF8}{min}サブカルチャーの特徴を決定付ける性質は、言語、美、宗教、政治、性、地理、またはそれらの要素の組み合わせかもしれません。\end{CJK} & \begin{CJK}{UTF8}{min}サブカルチャーを特徴的なものとする品質は、言語的、美学的、宗教的、政治的、性的、地理的、あるいはそれらの要因の組み合わせであることがあります。\end{CJK} & \begin{CJK}{UTF8}{min}そのような品質は、サブカルチャーを特徴的なものとするものである可能性があります。言語的、美学的、宗教的、政治的、性的、地理的、あるいはそれらの要因の組み合わせであることがあります。\end{CJK} \\ \midrule
    Japanese & New Kingdom ancient Egyptians marvelled at their predecessors monuments, which were then well over a thousand year old. & \begin{CJK}{UTF8}{min}新王国の古代エジプト人たちは、当時千年以上前に建てられた前身のモニュメントに驚嘆しました。\end{CJK} & \begin{CJK}{UTF8}{min}新王国時代のエジプト人は、当時すでに1000年以上も古い前人のモニュメントに驚嘆しました。\end{CJK} & \begin{CJK}{UTF8}{min}新王国時代のエジプト人は、それらの先人の墓苑が当時すでに1000年以上も古いことに驚嘆しました。\end{CJK} \\ \midrule
    Japanese & The U.S. Corps of Engineers estimated that 6 inches of rainfall could breach the previously damaged levees. & \begin{CJK}{UTF8}{min}米工兵隊は、以前に決壊した堤防が6インチの降水量で再び決壊する可能性があると推定しています。\end{CJK} & \begin{CJK}{UTF8}{min}米国エンジニア兵団は、6インチの降雨で以前に損傷した堤防が浸水すると推定しています。\end{CJK} & \begin{CJK}{UTF8}{min}以前に損傷を受けた堤防が、米国エンジニア兵団が推定する6インチの降雨で浸水するかもしれません。\end{CJK} \\ \midrule
    French & Although three people were inside the house when the car impacted it, none of them were hurt. & Bien que trois personnes aient été présentes dans la maison quand la voiture l'a percutée, aucune n'a été blessée. & Bien que trois personnes se trouvaient à l'intérieur de la maison lorsque la voiture a percuté celle-ci, aucune d'entre elles n'a été blessée. & Même si trois personnes étaient à l'intérieur de la maison lorsque la voiture a heurté celle-ci, aucune d'entre elles n'a été blessée. \\ \midrule
    French & 34 per cent of those in the poll share this view, wanting Queen Elizabeth II to be Australia's last monarch. & 34 \% des personnes interrogées partagent ce point de vue, et souhaitent que la reine Elizabeth II soit la dernière monarque à régner sur l'Australie. & 34 pour cent de ceux qui ont participé à l'enquête partagent cet avis, souhaitant que la reine Elizabeth II soit la dernière monarchie d'Australie & 34 pour cent de ceux qui ont participé à l'enquête souhaitent que la reine Elizabeth II soit la dernière monarchie d'Australie. \\ \midrule
    French & The qualities that determine a subculture as distinct may be linguistic, aesthetic, religious, political, sexual, geographical, or a combination of factors. & Les critères qui déterminent une sous-culture comme distincte peuvent être linguistiques, esthétiques, religieux, politiques, sexuels, géographiques ou une combinaison de facteurs. & Les qualités qui déterminent une sous-culture comme distincte peuvent être linguistiques, esthétiques, religieuses, politiques, sexuelles, géographiques ou une combinaison de facteurs." & Les caractéristiques qui déterminent une sous-culture comme distincte peuvent être linguistiques, esthétiques, religieuses, politiques, sexuelles, géographiques ou une combinaison de ces éléments. \\ \midrule
    French & New Kingdom ancient Egyptians marvelled at their predecessors monuments, which were then well over a thousand year old. & Les Égyptiens antiques du Nouvel Empire étaient émerveillés par les monuments de leurs prédécesseurs, qui avaient alors bien plus de mille ans. & Les Égyptiens anciens de la Nouvelle-République admiraient les monuments de leurs prédécesseurs, qui étaient alors bien plus âgés d'un millénaire. & Les Égyptiens anciens de la Nouvelle-République admiraient les monuments de leurs prédécesseurs, qui avaient alors plus de mille ans. \\ \midrule
    French & The U.S. Corps of Engineers estimated that 6 inches of rainfall could breach the previously damaged levees. & D’après le U.S. Corps of Engineers, une pluviométrie de 6 pouces pourrait ouvrir une brèche dans les digues précédemment endommagées. & Le Corps des ingénieurs des États-Unis a estimé qu'une pluie de 15 cm pourrait affaiblir les digues précédemment endommagées. & Le Corps des ingénieurs des États-Unis a estimé qu'une pluie de 15 cm pourrait mettre à mal les digues précédemment endommagées. \\ \midrule
    Indonesian & Although three people were inside the house when the car impacted it, none of them were hurt. & Walau ada tiga orang di dalam rumah yang ditabrak mobil, tak satu pun yang cedera. & Meskipun tiga orang ada di dalam rumah saat mobil menabraknya, tidak ada yang terluka. & Meskipun tiga orang ada di dalam rumah saat mobil menabraknya, mereka semua selamat. \\ \midrule
    Indonesian & 34 per cent of those in the poll share this view, wanting Queen Elizabeth II to be Australia's last monarch. & Dalam jajak pendapat mengenai pandangan ini, 34 persen dari mereka menginginkan Ratu Elizabeth II menjadi pemimpin monarki terakhir Australia. & 34 persen dari orang yang terlibat dalam polling ini memiliki pandangan yang sama, yaitu ingin Ratu Elizabeth II menjadi ratu terakhir di Australia. & 34 persen dari orang yang terlibat dalam polling ini memiliki pandangan yang sama, yaitu ingin Ratu Elizabeth II menjadi ratu terakhir di Australia. \\ \midrule
    Indonesian & The qualities that determine a subculture as distinct may be linguistic, aesthetic, religious, political, sexual, geographical, or a combination of factors. & Kualitas yang menentukan suatu subkultur berbeda dari yang lain bisa berupa bahasa, estetika, agama, politik, seksual, geografis, atau kombinasi sejumlah faktor. & Kualitas yang menentukan suatu subkultur sebagai distinct dapat berupa linguistik, estetika, agama, politik, seksual, geografis, atau kombinasi faktor-faktor tersebut. & Faktor-faktor yang menentukan sebuah subkultur sebagai distinct bisa berupa faktor linguistik, estetika, agama, politik, seksual, geografis, atau kombinasi dari beberapa faktor tersebut. \\ \midrule
    Indonesian & New Kingdom ancient Egyptians marvelled at their predecessors monuments, which were then well over a thousand year old. & Kerajaan Baru Mesir kuno mengagumi monumen para pendahulunya, yang saat itu berusia lebih dari seribu tahun. & Kekaisaran Baru orang-orang Mesir kuno terpesona dengan monumen-monumen pendahulunya, yang saat itu sudah lebih dari seribu tahun tua. & Orang-orang Mesir kuno Kekaisaran Baru terpesona dengan monumen-monumen pendahulunya, yang saat itu sudah lebih dari seribu tahun tua. \\ \midrule
    Indonesian & The U.S. Corps of Engineers estimated that 6 inches of rainfall could breach the previously damaged levees. & U.S. Corps of Engineers memperkirakan bahwa curah hujan sebesar 6 inci ini mampu menerobos tanggul yang sudah rusak. & Korps Engineer AS menghitung bahwa hujan 6 inci dapat merusak parit yang sebelumnya sudah rusak. & Badan Engineer AS memperkirakan bahwa hujan 6 inci dapat menembus parit yang sebelumnya sudah rusak. \\ \midrule
    Javanese & Although three people were inside the house when the car impacted it, none of them were hurt. & Sanadyan uwong telu ana ing njero omah nalika mobil kuwi nabrak, nanging kabeh selamet. & Walaupun tiga orang berada di dalam rumah saat mobil menabraknya, tidak ada di antara mereka yang terluka. & Walaupun ana ing tiga wong ing rumah kala mobil mbentak ing kono, ora ana saka ingkang terluka. \\ \midrule
    Javanese & 34 per cent of those in the poll share this view, wanting Queen Elizabeth II to be Australia's last monarch. & 34 per sen sing menehi swara sesawangane ngene, kepengin Ratu Elizabeth II dadi raja sing keri dhewe ana ning Australia. & Sebagian 34 persen dari mereka yang terlibat dalam polling ini berpendapat demikian, ingin Ratu Elizabeth II menjadi Ratu Australia terakhir. & Sebagian 34 persen dari mereka yang terlibat dalam polling ini memiliki pandangan yang sama, yaitu menginginkan Ratu Elizabeth II menjadi Ratu Australia terakhir. \\ \midrule
    Javanese & The qualities that determine a subculture as distinct may be linguistic, aesthetic, religious, political, sexual, geographical, or a combination of factors. & Kualitas sing nemtokake sawijining cabang kabudayan mbokmenawa basa, estetika, sing ana gayutane karo agama, politik, seksual, geografis, utawa sawijining kombinasi saka faktor-faktor. & Sifat-sifat yang menentukan subkultur sebagai yang terpisah bisa berupa linguistik, estetik, agama, politik, seksual, geografis, atau kombinasi faktor-faktor. & Kualitas-kualitas sing menentukan subkultur minangka distinct bisa ngandhut linguistik, estetik, agama, politik, seksual, geografis, utawa gabungan faktor-faktor. \\ \midrule
    Javanese & New Kingdom ancient Egyptians marvelled at their predecessors monuments, which were then well over a thousand year old. & Krajan Anyar Mesir kuno nggumun marang monumen-monumen leluhure, sing nalika iku tetep apik luwih seka sewu taun. & Rakyat Kerajaan Baru dari Mesir Kuno terpesona oleh monumen-monumen pendahulunya, yang saat itu sudah lebih dari seribu tahun tua. & Rakyat Kerajaan Baru dari Mesir Kuno terpesona oleh monumen-monumen pendahulunya, yang saat itu sudah lebih dari seribu tahun tua. \\ \midrule
    Javanese & The U.S. Corps of Engineers estimated that 6 inches of rainfall could breach the previously damaged levees. & Korps Insinyur AS ngira manawa udan 6 inci bisa nembus tanggul sing rusak sadurunge. & Korps Insinyur Amerika Serikat menghitung bahwa hujan 6 inci bisa merusak parit yang sebelumnya rusak. & Korps Insinyur Amerika Serikat menaksir bahwa hujan 6 inci bisa merusak parit yang telah rusak sebelumnya. \\ \midrule
    Sundanese & Although three people were inside the house when the car impacted it, none of them were hurt. & Sanaos tilu urang aya di jero bumi pas mobil na nabrak, teu aya anu cidera. & Walaupun tiga orang berada di dalam rumah saat mobil menabraknya, tidak ada yang terluka. & Walaupun ada tiga orang di dalam rumah saat mobil menabraknya, tak ada yang terluka. \\ \midrule
    Sundanese & 34 per cent of those in the poll share this view, wanting Queen Elizabeth II to be Australia's last monarch. & 34 persen jalma dina polling di gaduh pendapat ieu, nyaeta hoyong Ratu Elizabeth II janten monarkis pamungkas di Australia. & 34 persen dari mereka yang terlibat dalam survei memiliki pandangan ini, yaitu ingin Ratu Elizabeth II menjadi ratu terakhir Australia. & 34 persen dari mereka yang terlibat dalam survei memiliki pandangan yang sama, yaitu ingin Ratu Elizabeth II menjadi ratu terakhir di Australia. \\ \midrule
    Sundanese & The qualities that determine a subculture as distinct may be linguistic, aesthetic, religious, political, sexual, geographical, or a combination of factors. & Kualitas anu nangtukeun subkultur salaku bédana panginten basa, éstétis, religius, politik, séks, géografis, atanapi kombinasi faktor. & Sifat-sifat yang menentukan subkultur sebagai terpisah mungkin bersifat linguistik, estetika, agama, politik, seksual, geografis, atau kombinasi dari faktor-faktor tersebut. & Sifat-sifat nu nentukkeun subkultur salaku nu béda bisa jadi linguistik, estetika, agama, politik, seksual, geografis, atawa kombinasi ti faktor-faktor ieu. \\ \midrule
    Sundanese & New Kingdom ancient Egyptians marvelled at their predecessors monuments, which were then well over a thousand year old. & Karajaan anyar urang Mesir kuno takjub kana monumén paninggalan karuhunna, anu umurna langkung ti sarebu taun. & Para raja anyar dina Egypt kuno nyogog nara-nara jaman awal, anu dina jaman ayeuna kira-kira ribuan taun. & Para raja anyar dina Egypt kuno terkagum-kagum kana nara-nara candi jaman awal, anu dina jaman ayeuna kira-kira leuweung taun. \\ \midrule
    Sundanese & The U.S. Corps of Engineers estimated that 6 inches of rainfall could breach the previously damaged levees. & Korps insinyur A.S ngira-ngira curah hujan 6 inci tiasa nembus tanggul anu samemehna geus ruksak. & Korps Engineer AS diperkirakan bahwa 6 inci hujan bisa merusak benteng yang sebelumnya sudah rusak. & Korps Engineer AS diperkirakan bahwa hujan sebesar 6 inci dapat merusak benteng yang sudah rusak sebelumnya.\\
    \bottomrule    
    \caption{Examples of ChatGPT translated and post-edited sentences.}
    \label{tab:my_label}
\end{longtable}
\end{landscape}

% \section{Evaluation Results for Reasoning}
% \label{sec:reasoning-result}

% We provide the complete results for reasoning tasks on~\Cref{tab:reasoning}.
% \begin{table}[]
% \resizebox{\linewidth}{!}{%
% \centering
% \begin{tabular}{ccc}
% \toprule
% \textbf{Categories} & \textbf{Testset} & \textbf{Result} \\ \midrule
% Deductive & \begin{tabular}[c]{@{}c@{}}EntailmentBank\\ bAbI (task 15)\end{tabular} & \begin{tabular}[c]{@{}c@{}}28/30\\ 28/30 (as-is: 19/30)\end{tabular} \\\midrule
% Inductive & \begin{tabular}[c]{@{}c@{}}CLUTRR\\ bAbI (task16)\end{tabular} & \begin{tabular}[c]{@{}c@{}}13/30\\ 20/30 (as-is: 0/30)\end{tabular} \\\midrule
% Abductive & $\alpha$NLI & 26/30 \\\midrule
% Mathematical & Math & 13/30 \\\midrule
% Temporal & Timedial & 26/30 \\\midrule
% Spatial & \begin{tabular}[c]{@{}c@{}}SpartQA (hard)\\SpartQA (basic)\\ 
% StepGame (hard)\\ StepGame (basic)\\  StepGame (cardinal)\\ StepGame (diagonal)\\ StepGame (clock)\end{tabular} & \begin{tabular}[c]{@{}c@{}}8/32\\20/32\\ 7/30\\ 19/30\\  17/20\\ 11/20\\ 5/20\end{tabular} \\\midrule
% Commonsense & \begin{tabular}[c]{@{}c@{}}CommonsenseQA\\ PIQA\\ Pep-3k (Hard)\end{tabular} & \begin{tabular}[c]{@{}c@{}}27/30\\  25/30\\ 28/30\end{tabular} \\\midrule
% Causal & E-Care & 24/30 \\\midrule
% Multi-hop & hotpotQA & 8/30 \\\midrule
% Analogical & Letter string analogy & 30/30 \\ \bottomrule
% \end{tabular}}
% \caption{Composed results for all reasoning tasks.}
% \label{tab:reasoning}
% \end{table}

\clearpage
\twocolumn

\end{document}